\documentclass{article}
\usepackage{amssymb}

\usepackage{xcolor}
\usepackage{amsmath}
\usepackage{xspace}
\usepackage{dsfont}
\usepackage{graphicx}
\usepackage{tabularx}
\usepackage{multirow}
\usepackage{booktabs}
\usepackage{enumitem}
\usepackage{longtable} 
\usepackage{pdfpages}

\usepackage{listings}
\usepackage{xcolor}
\usepackage[font=footnotesize]{caption}

\usepackage{amsmath,amsfonts,bm}

\def\figref#1{Figure~\ref{#1}}
\def\Figref#1{Figure~\ref{#1}}

\def\Tabref#1{Table~\ref{#1}}

\def\eqref#1{equation~\ref{#1}}

\def\1{\bm{1}}

\def\vg{{\bm{g}}}

\def\vp{{\bm{p}}}

\def\vs{{\bm{s}}}

\DeclareMathAlphabet{\mathsfit}{\encodingdefault}{\sfdefault}{m}{sl}
\SetMathAlphabet{\mathsfit}{bold}{\encodingdefault}{\sfdefault}{bx}{n}

\def\gG{{\mathcal{G}}}

\newcommand{\shortpara}{\noindent\textbf}

\newcommand{\model}{StackItUp\xspace}
\newcommand{\objectSet}{\mathcal{O}\xspace}
\newcommand{\objectTypeSet}{T\xspace}
\newcommand{\sketch}{\mathcal{S}\xspace}

\newcommand{\typeSet}{\mathcal{T}\xspace}
\newcommand{\poseSet}{\mathcal{P}\xspace}
\newcommand{\noise}{\ensuremath{\epsilon}}

\newcolumntype{P}[1]{>{\centering\arraybackslash}p{#1}}
\newcolumntype{M}[1]{>{\centering\arraybackslash}m{#1}}

\newcommand{\tfMatrix}{\mathbf{H}}
\newcommand{\goalPose}{\tfMatrix^{\mathrm{goal}}}
\newcommand{\initPose}{\tfMatrix^{\mathrm{init}}}
\newcommand{\graspPose}{\tfMatrix^{\mathrm{grasp}}}
\newcommand{\setTfMatrix}{\mathcal{H}}
\newcommand{\setTfMatrixPick}{\setTfMatrix^{\mathrm{pick}}}
\newcommand{\setTfMatrixPrePlace}{\setTfMatrix^{\mathrm{pre-place}}}
\newcommand{\setTfMatrixPostPlace}{\setTfMatrix^{\mathrm{post-place}}}

\makeatletter
\newcommand{\onedot}{\ifx\@let@token.\else.\null\fi}
\makeatother
\def\eg{\emph{e.g}\onedot} 
\def\ie{\emph{i.e}\onedot} 
 
 \def\vs{\emph{vs}\onedot}

\makeatother

\usepackage[final]{corl_2025} 

\title{``Stack It Up!'': 3D Stable Structure Generation from 2D Hand-drawn Sketch}

\author{
Yiqing Xu\textsuperscript{1}\thanks{\texttt{xuyiqing@comp.nus.edu.sg}} \quad
Linfeng Li\textsuperscript{1} \quad
Cunjun Yu\textsuperscript{1} \quad
David Hsu\textsuperscript{1,2} \\
\textsuperscript{1}School of Computing, \\
\textsuperscript{2} Smart Systems Institute,\\
National University of Singapore
}

\begin{document}
\maketitle

\begin{figure}[h]
\vspace{-25pt}
\centering\includegraphics[width=0.98\linewidth]{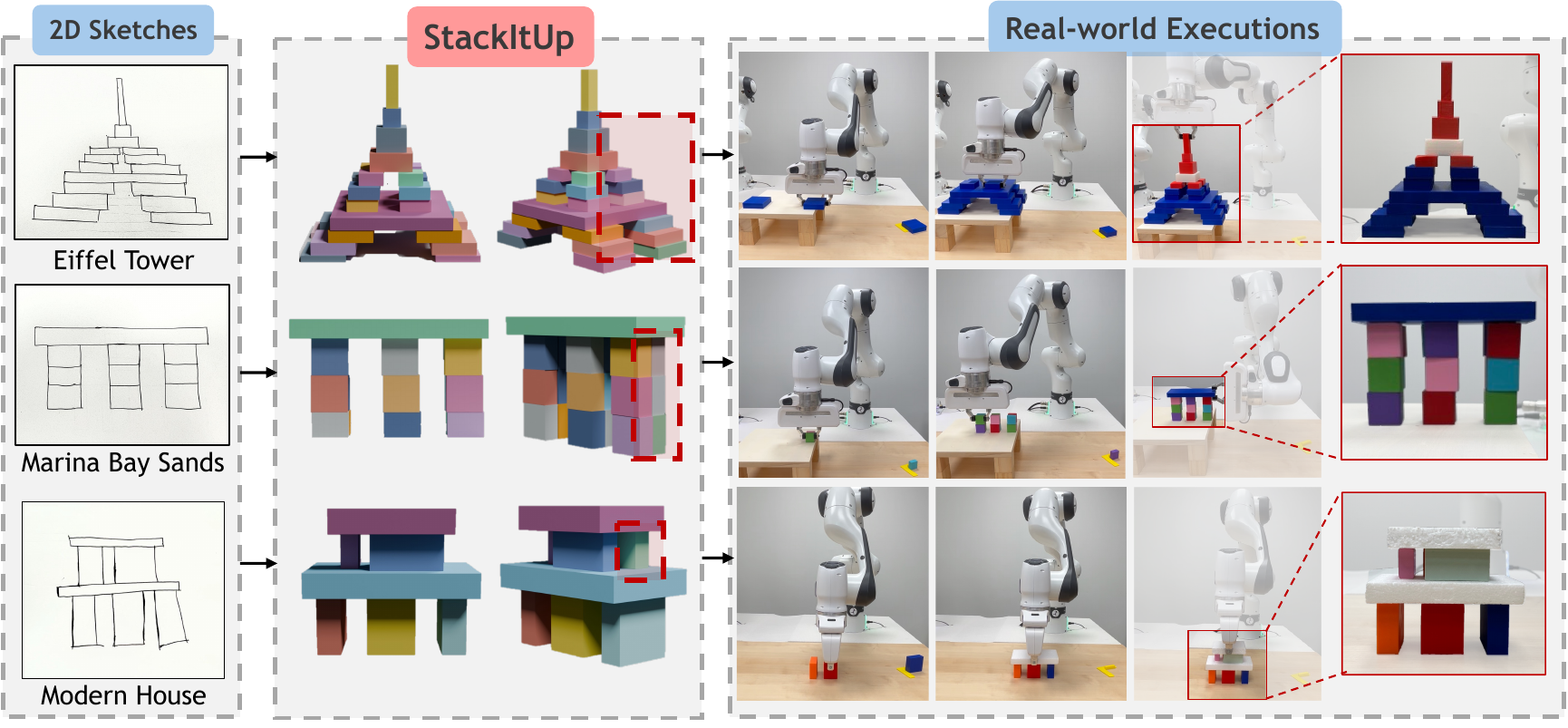}
\caption[Demonstration of StackItUp: Generating Multi-level 3D Stable Structure from 2D Rough Hand-Drawn Sketches.]{\textbf{Demonstration of \model.} \model\ allows non-experts to specify 3D structure for robot execution using a simple 2D sketch. From a rough front-view drawing, it predicts accurate 3D poses and hidden supports to generate stable structures that resemble the sketch. These poses can be directly used by a robotic arm as goal specification for motion planning and physical execution. Shown: input sketches (left), generated 3D structures with predicted supports highlighted (middle), and real robot executions (right).}
    \label{fig:teaser}
\vspace{-15pt}
\end{figure}

\begin{abstract}
Imagine a child sketching the Eiffel Tower and asking a robot to bring it to life. Today’s robot manipulation systems can’t act on such sketches directly—they require precise 3D block poses as goals, which in turn demand structural analysis and expert tools like CAD. We present \emph{\model}, a system that enables non-experts to specify complex 3D structures using only 2D front-view hand-drawn sketches. \model\ introduces an abstract relation graph to bridge the gap between rough sketches and accurate 3D block arrangements, capturing the symbolic geometric relations (\eg, \emph{left-of}) and stability patterns (\eg, \emph{two-pillar-bridge}) while discarding noisy metric details from sketches. It then grounds this graph to 3D poses using compositional diffusion models and iteratively updates it by predicting hidden internal and rear supports—critical for stability but absent from the sketch. Evaluated on sketches of iconic landmarks and modern house designs, \model\ consistently produces stable, multilevel 3D structures and outperforms all baselines in both stability and visual resemblance.
\end{abstract}
\vspace{-15pt}


\section{Introduction}
\vspace{-7pt}
Imagine a child drawing a simple sketch of Eiffel Tower and asking a robot to build it. A human grasps the idea at a glance, but the robot freezes: the current robotic system requires the manipulation goal to be fully specified as exact 3D object poses—typically crafted in design software and verified by force analysis—before motion planning and execution can even begin~\citep{simeon2002manipulation,6631099,suarez-ruiz2016framework,zucker2013chomp,karaman2011sampling,kavraki1996probabilistic, lavalle1998rapidly}. This creates a significant barrier for non-experts, making robot systems inaccessible to everyday users.

We propose \emph{\model}, a system that enables non-experts to specify complex 3D structures for robot manipulation using only a single 2D front-view hand-drawn sketch (\figref{fig:teaser}). The challenge lies in bridging the gap between the rough drawing and a complete 3D goal specification. First, hand-drawn sketches are metrically imprecise—objects may be distorted, misaligned, or physically implausible, making direct pose recovery unreliable. Second, a front-view sketch—precisely because it avoids perspective—is inherently incomplete: it omits interior and rear supports that are crucial for physical stability~\cite{babivc2016culture,hunt2013bim, novelbim2023problems, marsbim2023drawings, tekla2023harming,shalin2023modeling}. To construct such a stable structure, the robot must infer both the missing supporting elements and the metrically accurate 3D poses of all blocks for execution.

Jointly predicting missing blocks and precise 3D poses creates a vast discrete–continuous search space ~\cite{10160365,irshad2022shapo,garrett2018ffrob,garrett2018sampling}. Our key idea is to introduce an \emph{abstract relation graph} as an intermediate representation to manage this complexity \citep{yang2023diffusion, Xu-RSS-24, zhu2021hierarchical}. This graph captures high-level geometric relations (\eg, \emph{left-of}) and stability patterns (\eg, \emph{two-pillar-bridge}), distilling essential structural cues while discarding noisy metric details from the sketch. When hidden supports are required for stability, subgraph matching against a dictionary of stability patterns efficiently reveals likely missing blocks and their relations. Moreover, this symbolic abstraction also enables flexible and scalable 3D pose generation: given any graph, we compose diffusion-based pose generators trained on individual relation types, and perform MCMC sampling over their combined scores to generate a coherent, stable 3D arrangement~\cite{yang2023diffusion,Xu-RSS-24,du2023reduce,sjoberg2023mcmc}.

\model\ operates in two stages, as shown in \figref{fig:overview}. First, it extracts an abstract relation graph from the sketch, capturing geometric relations and stability patterns among visible objects. Then, as illustrated in \figref{fig:iterative}, it grounds the graph to 3D poses using compositional diffusion models, checks stability via simulation, and iteratively updates this graph by adding hidden supports and relations as guided by the stability patterns until the generated 3D arrangement is stable.

We evaluate \model\ on a diverse set of hand-drawn sketches depicting historical buildings (\eg, Taj Mahal), iconic landmarks (\eg, Marina Bay Sands), and modern architectural designs. As shown in \figref{fig:demo_instances}, these examples vary widely in appearance, internal support structure, and block dimensions. In all cases, \model\ produces physically stable, multilevel 3D arrangements that faithfully reflect the sketch intent and outperform all baselines in both stability and visual resemblance.

\section{Related Works}
\vspace{-7pt}
\noindent\textbf{Sketch-Based Goal Specification for Robotics.} Sketches have long been used in robotic systems to specify goals such as navigation targets, obstacles, or object interactions~\citep{barber2010sketch, porfirio2023sketching}. More recent work explores sketch-based inputs for high-level manipulation tasks. For example, RT-Sketch~\citep{sundaresan2024rt} uses hand-drawn inputs to guide object rearrangement, while others~\citep{cui2022can, gu2023robotic} use sketches as part of multimodal interfaces for task planning. However, these methods primarily focus on scene-level or trajectory goals, and do not address structure-centric reasoning. In contrast, \model\ targets a fundamentally different challenge: generating physically stable, multi-object 3D structures from rough 2D sketches, which requires inferring both spatial layout and unseen supports.\\
\noindent\textbf{User-Friendly Input for 3D Structure Generation.}
Recent systems translate high-level inputs such as natural language into object-centric structures. Blox-Net~\citep{goldberg2024blox} leverages vision-language models to generate 3D assemblies that are physically plausible and robot-executable. QUERY2CAD~\citep{badagabettu2024query2cad} refines CAD models through natural language corrections. While these methods highlight the potential of user-friendly inputs for structure generation, they lack fine-grained spatial control and rely on natural language’s limited ability to express geometric detail. In contrast, our sketch-based approach captures fine-grained spatial details and integrates explicit stability reasoning, thereby addressing a gap left by natural language–based methods. \\
\noindent\textbf{3D Structure Generation from Single-View 2D Inputs.}
A related direction involves generating 3D object structures from single-view 2D inputs~\cite{hong2024lrm, tang2024dreamgaussian}. StackGen~\citep{sun2024stackgen} uses diffusion models to produce stacking layouts from silhouettes. However, it operates in 2D ($x$–$z$ plane) and optimizes for visual similarity, without explicit reasoning about physical stability. Other methods such as Part123~\citep{liu2024part123} and follow-ups~\citep{liu2023zero, liu2023one, liu2024one} reconstruct metrically accurate 3D models from images. Yet, these methods typically require photorealistic input views and produce unified meshes, making them unsuitable for block-based robot stacking tasks. Moreover, their reliance on fully observed objects limits applicability when key structural elements are occluded or missing.

\vspace{-7pt}
\section{Problem Formulation}
\vspace{-8pt}
\model enables non-experts to specify their desired multi-level 3D structures to robots by drawing a 2D rough sketch, as shown in \Figref{fig:teaser}. This sketch $\sketch$ is a front-view illustration of the desired structure that shows a set of visible blocks with labeled types. From this sketch, \model\ must choose a set of blocks from a given library $\objectTypeSet=\{\tau_1,\dots,\tau_N\}$—each type $\tau_i$ representing a rigid, axis-aligned block of size $g_i=(w_i,l_i,h_i)$—and assign metric poses so that the resulting structure (i) remains static under gravity and (ii) closely resembles the user's sketch. Formally, the output is a set of blocks
$\objectSet=\{o_1,\dots,o_M\}$ with
$o_i=(\tau_i,p_i)$, where $\tau_i\in\objectTypeSet$ is the block type and
$p_i=(x_i,y_i,z_i)$ is the block’s centroid. Equivalently, we write
$\objectSet=(\typeSet,\poseSet)$, with
$\typeSet=(\tau_1,\dots,\tau_M)$ collects the block types and 
$\poseSet = (p_1, \dots, p_M)$ for the corresponding poses. The sketch provides type labels for visible blocks $\typeSet_{\text{obs}}$ but inaccurate pose cues $\poseSet_{\text{obs}}$; it may also omit interior or rear supports.  \model\ must therefore decide whether additional hidden blocks
$\objectSet_{\text{hid}}$ are required, and predict precise 3D poses $\poseSet$ for \emph{all} blocks. The final goal specification
$\objectSet=\objectSet_{\text{obs}}\cup\objectSet_{\text{hid}}$
must be physically stable and, when projected to the front view, preserve the abstract geometric relations implied by~$\sketch$. 

\vspace{-7pt}
\section{Stack It Up}
\vspace{-7pt}

\begin{figure*}[t]
    \centering    \includegraphics[width=0.9\linewidth]{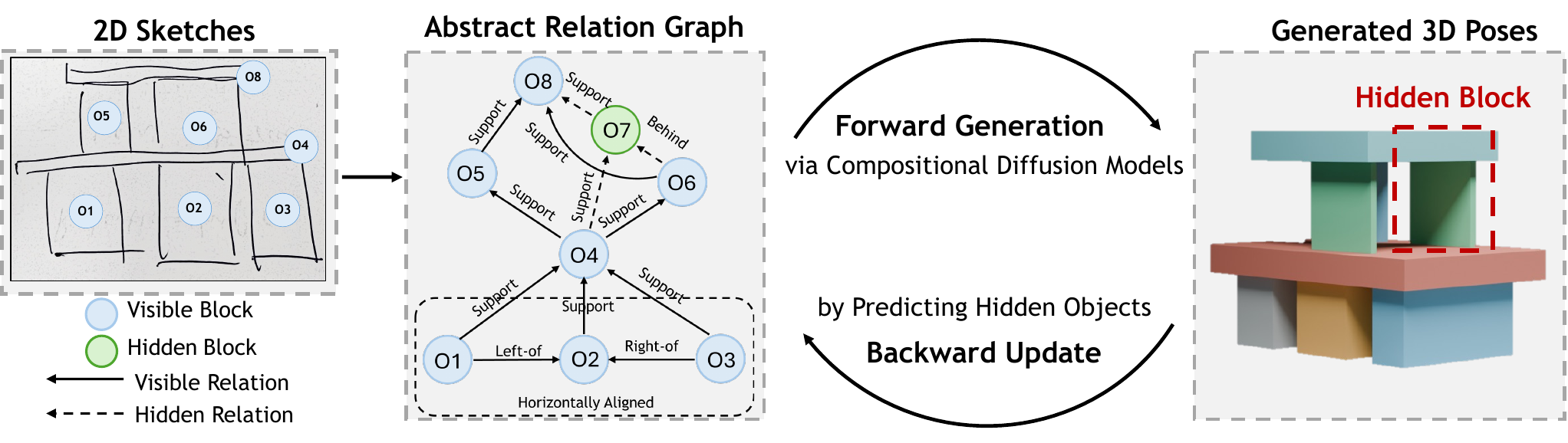}
    \vspace{-10pt}
    \caption[StackItUp Method Overview.]{\textbf{Method overview.} \model\ uses an abstract relation graph $\gG$ as an intermediate between a rough 2D sketch (left) and the generated 3D block arrangement (right). The graph (middle) encodes high-level geometric and stability relations while abstracting away exact poses. \model\ first extracts $\gG_0$ from visible blocks in the sketch (blue nodes), then iteratively grounds it to 3D poses using compositional diffusion models. If instability is detected, the graph is updated with predicted hidden supports (green nodes), and re-grounded to 3D poses.} 
    \label{fig:overview}
    \vspace{-15pt}
\end{figure*}

\model transforms a metrically imprecise 2D front-view sketch~$\sketch$ into a 3D block arrangement~$\objectSet$ that is both physically stable and visually resembles the sketch. Central to \model is an abstract relation graph $\gG$, where nodes represent blocks and edges capture qualitative geometric relations (\eg, \emph{left-of}) and local stability patterns (\eg, \emph{two-pillar-bridge}), as an intermediate representation between the 2D sketch $\sketch$ and the 3D block arrangement $\objectSet$ (see \Tabref{tab:relationships} for a complete list of relations used in this work). With this abstraction, the task splits naturally into two phases (\Figref{fig:overview}):
\begin{enumerate}[label=\textbf{\arabic*.},  
                leftmargin=1.5em,
                itemsep=0.em,
                topsep=-3pt,
                partopsep=0pt]  
    \item \textbf{One-time Graph Extraction.} Extract an initial graph $\gG_0$ from sketch~$\sketch$ for the visible blocks.
    \item \textbf{Iterative Graph Grounding.} Ground the graph $\gG_t$ to 3D poses $\poseSet_t$ through compositional diffusion models, test them in simulation, and—when instability is detected—extend the graph with hidden supports, \ie, $\gG_{t+1} = \gG_{t} \cup \gG_{t}^{\text{hid}}$. The loop terminates once the structure stands.
\end{enumerate}
Section~\ref{ssec:abstract_relation_graph} presents the geometric relation and stability pattern libraries and describes the one-time graph extraction from the sketch; Section~\ref{ssec:iterative_pose_generation} outlines the iterative forward–backward grounding algorithm that alternates between pose generation and graph update.

\begin{table}[tp]
    \centering
    \caption{ Library of abstract relations, including both geometric relations and stability patterns.}
    \label{tab:relationships}
    \scriptsize
    \setlength{\tabcolsep}{2pt}
    \renewcommand{\arraystretch}{1.1} 
\begin{tabular}{p{0.23\linewidth} p{0.25\linewidth} p{0.25\linewidth} p{0.23\linewidth}}
\toprule
\multicolumn{4}{l}{\textbf{Geometric Relations}}                                                                     \\ \hline
\multicolumn{2}{l}{\textbf{Front View, $x$-$z$ plane}}   & \multicolumn{2}{l}{\textbf{Top-down View, $x$-$y$ plane}} \\
\cmidrule(lr){1-2} \cmidrule(lr){3-4} 
left-of                      & left-in                   & front-of                     & front-in                   \\
right-in                     & center-in                 & back-in                      & touching-along-y           \\
supported-by-partially       & supported-by-fully        & near-along-y                 & depth-aligned              \\
horizontal-aligned           & vertical-aligned-centroid & depth-aligned-in-a-line      & regular-grid-sparse        \\
vertical-aligned-left        & vertical-aligned-right    & regular-grid-compact         & random-split-grid-sparse   \\
horizontal-aligned-in-a-line & touching-along-x          & random-split-grid-compact    &                            \\
near-along-x                 &                           &                              &                            \\
\toprule
\multicolumn{4}{l}{\textbf{Stability Patterns}}                                                                      \\ \hline
single-block-stack  & cantilever-with-counterbalance & two-pillar-single-top-bridge & n-pillar-single-top-bridge  \\
single-base-n-pillar-bridge &  two-base-single-overhead-pyramid     & n-base-single-overhead-pyramid
& single-base-n-overhead-pyramid \\
n-base-m-overhead-pyramid & basic-arc \\
\bottomrule
\end{tabular}
\vspace{-20pt}
\end{table}

\vspace{-7pt}
\subsection{Abstract Relation Graph Extraction}\label{ssec:abstract_relation_graph}
\vspace{-7pt}
We use an abstract relation graph $\gG$ as an intermediate representation between a 2D sketch $\sketch$ and the final 3D block arrangement $\objectSet$. This graph compactly encodes the qualitative geometric and stability relations that define the structure of multi-layer 3D structures. Formally, $\gG$ consists of nodes representing blocks with their types $\typeSet$, and edges denoting abstract relations from a set $\mathcal{R}$. Each relation $r_i(o^i_1, \ldots, o^i_{k_i}; R_i) \in \gG$ has type $R_i \in \mathcal{R}$ and arity $k_i$, applying to a subset of blocks $\{o^i_1, \ldots, o^i_{k_i}\}$. The poses $\{p^i_1, \ldots, p^i_{k_i}\}$ are abstracted away by this intermediate representation. We divide relation types into geometric relations $\mathcal{R}_{\text{geom}}$ (e.g., \emph{left-of}) that capture spatial layout, and stability patterns $\mathcal{R}_{\text{stab}}$ (e.g., \emph{two-pillar-bridge}) that encode physical support. We begin by defining these relations and then describe how to extract the initial graph $\gG_0$ from the sketch $\sketch$. \\
\noindent\textbf{Geometric relations.} We define $24$ qualitative geometric relations (top of \Tabref{tab:relationships}).
Among these, $13$ arise directly from the sketch because they depend only on bounding-box coordinates in the front ($x$–$z$) plane(\eg, \emph{left-of}, \emph{touching-along-x}). The remaining $11$ involve depths (\eg, \emph{front-of}, \emph{touching-along-y}), therefore requiring $x$–$y$ information and are only revealed once hidden supports are added.
Each relation $R\in\mathcal{R}_{\text{geom}}$ is detected by a rule-based classifier $h_R$ that inspects the bounding boxes projected into their respective planes; full rule sets appear in Appendix \ref{appendix:geometric_relation}. \\
\noindent\textbf{Local Stability Patterns.}
Beyond geometric relations, we model $10$ local stability patterns (bottom of \Tabref{tab:relationships}), each pattern $R \in \mathcal{R}_{\text{stab}}$ describes a class of stable two-layer subgraphs in which a base tier supports a top tier; arbitrary compositions of these patterns yield stable multi-level structures. Each stability pattern $R \in \mathcal{R}_{\text{stab}}$ has a classifier $h_{R}$ defined by four descriptors: (i) admissible counts $(\{n_{\text{base}}\}, \{n_{\text{top}}\})_{R}$ of blocks per tier; (ii) allowed ``\emph{supported-by}" subgraphs $\{\gG^{R}_{\text{supp}}\}$ connecting base to top blocks; permissible geometric subgraphs (iii) among base blocks $\{\gG^{R}_{\text{base}}\}$ and (iv) among top blocks $\{\gG^{R}_{\text{top}}\}$, including relations such as \emph{horizontally-aligned} and \emph{regular-grid}. Given a candidate graph $\gG^{\text{geom}}$ and a partition into top and base blocks, the classifier evaluation $h_R(o_{\text{base}}^{1}, \ldots, o_{\text{base}}^{n}, o_{\text{top}}^{1}, \ldots, o_{\text{top}}^{m})$ returns $1$ when this candidate graph matches an permissible instance for each classifier descriptor. Complete classifier specifications appear in Appendix \ref{appendix:stability_pattern}. \\
\noindent\textbf{Extracting Abstract Relation Graph from Sketch.}
The sketch~$\sketch$ depicts the desired 3D structure through enclosed boxes in a front ($x$–$z$) plane, where each box corresponds to a visible block $o_i \in \objectSet_{\text{obs}} = \{o_1, o_2, \dots, o_K\}$ with an labeled type $\tau_i \in \typeSet_{\text{obs}} = \{\tau_1, \tau_2, \dots, \tau_K\}$.  To handle imperfect or roughly drawn sketches, we first regularize the strokes by flood-fill and dilation, fitting an axis-aligned bounding box to each block and recording coarse dimensions $(\hat w_i,\hat h_i)$ and centroids $(\hat x_i,\hat z_i)$; these values only reflect qualitative geometric relations, but not the metrically accurate 3D pose $p_i=(x_i,y_i,z_i)$.  A set of rule-based geometric classifiers $h_R,\;R\!\in\!\mathcal R_{\text{geom}}$, then operates on the bounding boxes ${(\hat{w}_i, \hat{h}_i, \hat{x}_i, \hat{z}_i)}_{i=1}^K$ to populate the geometric graph $\gG_0^{\text{geom}}$.  Next, stability-pattern classifiers $h_R,\;R\!\in\!\mathcal R_{\text{stab}}$, search $\gG_0^{\text{geom}}$ for subgraph matches to known local stability patterns, producing the stability graph $\gG_0^{\text{stab}}$.  Their union yields the initial graph $\gG_0=\gG_0^{\text{geom}}\cup\gG_0^{\text{stab}}$.

\vspace{-7pt}
\subsection{Abstract Relation Graph Grounding}\label{ssec:iterative_pose_generation}
\begin{figure}
    \centering    \includegraphics[width=0.98\linewidth]{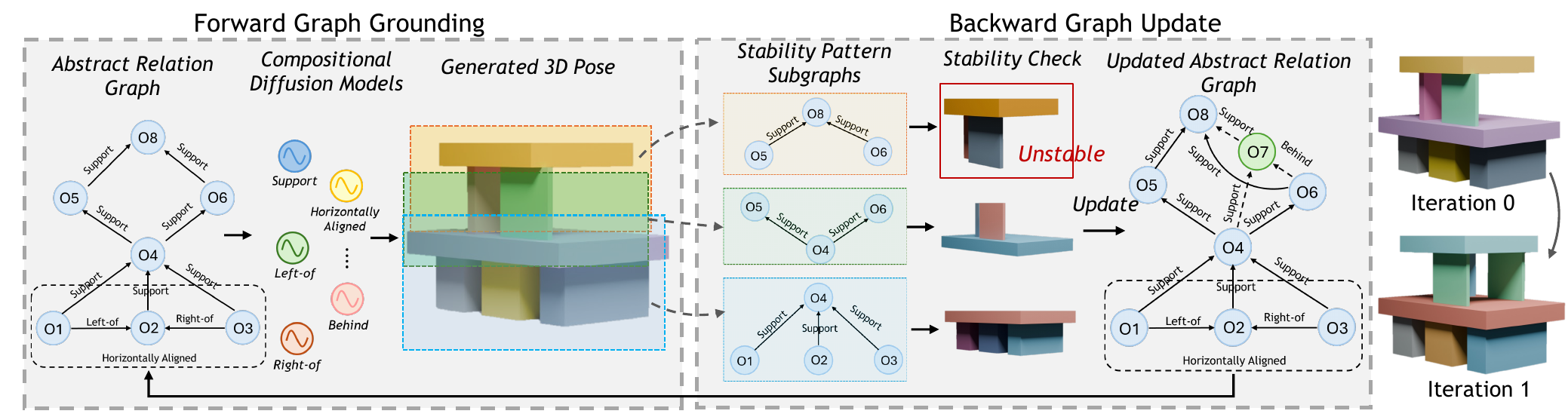}
    \vspace{-5pt}
    \caption[Illustration for Iterative Graph Grounding.]{\textbf{Iterative graph grounding.} In the forward step (left), given a graph $\gG_t$, compositional diffusion models generate 3D block poses. In the backward step (center-left), $\gG_t$ is decomposed into subgraphs based on stability patterns, and each is checked via physics simulation. If unstable, new support blocks (green nodes) and relations are added. These updates are aggregated into an updated graph $\gG_{t+1}$ for re-grounding. The right column shows the 3D structure before and after the graph update, with an added support ensuring stability. }
    \label{fig:iterative}
    \vspace{-15pt}
\end{figure}

\vspace{-7pt}
To ground the graph $\gG_0$ into stable 3D poses $\poseSet$, we avoid training a single monolithic model conditioned on the entire graph and instead use a set of specialized diffusion models $\{f_{R_0}, \ldots, f_{R_M}\}$, each trained on a relation type $R_i \in \mathcal{R}$. By composing these models based on the relations in $\gG_0 = \{r_0, \ldots, r_N\}$, we jointly predict initial poses $\poseSet_0$ for all blocks. However, since $\gG_0$—extracted from a front-view sketch—often omits hidden supports needed for stability, we introduce an iterative forward–backward grounding procedure (see \Figref{fig:iterative}). This process alternates between generating poses (forward) and augmenting the graph with missing supports when instability is detected (backward). The remainder of this section first describes the pre-training of relation-specific diffusion models, then details the iterative grounding algorithm.

\vspace{-7pt}
\paragraph{Training Individual Diffusion Models for Each Abstract Relation.} We train a set of specialized diffusion models $\{f_{R_0}, \ldots, f_{R_M}\}$, where each $f_R$ corresponds to an abstract relation $R \in \mathcal{R}_{\text{geom}} \cup \mathcal{R}_{\text{stab}}$. Each relation $R$ is defined by a rule-based classifier $h_R$ that verifies whether a tuple of block poses $\{p_1, \ldots, p_k\}$ satisfies the relation given their geometries $\{g_1, \ldots, g_k\}$. Its diffusion model $f_R$ learns to generate such valid poses by sampling from $q_R(\vp \mid \vg) \propto \mathds{1}[h_R(\vg, \vp)]$, where $\vp$ and $\vg$ are vectorized pose and geometry inputs, and $\mathds{1}[\cdot]$ is an indicator function enforcing relation satisfaction. To train each $f_R$, we generate synthetic data of multi-level 3D structures by composing multiple stability patterns. These structures exhibit diverse instances of both geometric and stability relations. We then apply the classifier ${h_{R}}$ to extract positive examples for each relation and construct datasets ${\mathcal{D}_{R}}$ accordingly. For each $R$, we construct a denoising diffusion model $f_R$~\citep{ho2020denoising} where the distribution $q_R(\vp \mid \vg)$ maximizes the likelihood of its dataset. Specifically, for a given sample $(\vg, \vp) \in \mathcal{D}_R$, we apply Gaussian noise to $\vp$ across $T$ time steps and train a denoising network $\epsilon_R(\vp_t, \vg, t)$ to recover the original poses. The training loss minimizes the following error: \begin{equation} \mathcal{L}_{\text{MSE}} = \mathbb{E}_{(\vg, \vp), \boldsymbol{\epsilon}, t} \left[ \left| \boldsymbol{\epsilon} - \epsilon_R\left(\sqrt{\bar{\alpha}_t}\vp + \sqrt{1 - \bar{\alpha}_t} \boldsymbol{\epsilon}, \vg, t\right) \right|^2 \right], \end{equation} where $\boldsymbol{\epsilon} \sim \mathcal{N}(\mathbf{0}, \mathbf{I})$, $t \sim \mathcal{U}(1, T)$, and $\bar{\alpha}_t$ is the diffusion denoising schedule. Sampling from $f_R$ involves reversing the diffusion process using the learned denoiser. Starting from $\vp_T \sim \mathcal{N}(\mathbf{0}, \mathbf{I})$, we apply a learned reverse kernel parameterized by $\epsilon_R$ to iteratively obtain $\vp_{t-1}$ from $\vp_t$ until reaching $\vp_0 \sim q_R(\cdot \mid \vg)$, a valid pose that satisfies $h_R$. See Appendix~\ref{appendix:DM_implementation} for implementation details.

\vspace{-7pt}
\paragraph{Iterative Pose Generation.} 
We alternate between grounding a graph into 3D poses and updating the graph to recover missing supports. This forward–backward loop combines compositional diffusion models for pose generation with stability-pattern-guided heuristics for efficient graph update.

\textbf{\emph{Forward Graph Grounding via Compositional Diffusion Models.}}
Given a $\gG$, we aim to sample valid 3D poses $\poseSet$ that satisfy all relations in the graph. Each edge $r \in \gG$ corresponds to a relation $R$ applied to a specific block set and is associated with a diffusion model $f_R$ that generates poses by approximating a conditional distribution $q_R(\vp^r \mid \vg^r)$ over the poses $\vp^r$ with geometries $\vg^r$. To predict poses for the entire graph $\gG$, we compose these models into a product distribution: $q_{\text{prod}}^t(\vp \mid \vg) := \prod_{r \in \gG} q_R^t(\vp^r \mid \vg^r),$ where each $q_R^t(\cdot)$ is the noised distribution at diffusion step $t$ for the relation $R$ associated with edge $r$. Our target is to sample from a \textbf{sequence} of product distributions $\{q_{\text{prod}}^t(\vp \mid \vg)\}_{t=0:T}$ starting from an initial sample $\vp_T$ drawn from $\mathcal{N}(\mathbf{0}, \mathbf{I})$. However, we cannot directly access the reverse diffusion kernel or exact score function for $q_{\text{prod}}^t(\vp \mid \vg)$, since it aggregates multiple models~\citep{du2023reduce, yang2023diffusion}. To approximate the transition from $q_{\text{prod}}^t(\vp \mid \vg)$ to $q_{\text{prod}}^{t-1}(\vp \mid \vg)$, we adopt a form of annealed MCMC that approximates the composite score function by summing the individual score functions from each model: $\nabla_{\vp_t} \log q_{\text{prod}}^t(\vp_t \mid \vg) \approx \sum_{r \in \gG} \noise_{R}(\vp_t^r, \vg^r, t),$
where $\noise_R$ is the learned denoising function for relation $R$. We apply the unadjusted Langevin algorithm (ULA) to perform one reverse diffusion step using the composite score function over the graph $\gG$: \begin{equation}\small \vp_{t-1} = \vp_t - A_t \sum_{r \in \gG} \noise_R(\vp_t^r, \vg^r, t) + B_t \boldsymbol{\xi}, \quad \boldsymbol{\xi} \sim \mathcal{N}(\mathbf{0}, \mathbf{I}), \end{equation} where $A_t$ and $B_t$ are constants from the diffusion schedule.  At each noise level $t$, we run $M$ iterations of ULA sampling using the composite score for the product distribution $q_{\text{prod}}^t(\cdot)$. Starting from $\vp_T \sim \mathcal{N}(\mathbf{0}, \mathbf{I})$, this iterative process gradually refines the sample, producing a final pose $\poseSet$ drawn from $q_{\text{prod}}^0(\cdot)$ that satisfies all relations in $\gG$.

\textbf{\emph{Backward Graph Update Guided by Stability Patterns.}}
The initial grounded poses $\poseSet_0$ from $\gG_0$ may be unstable due to missing hidden supports. Extending $\gG_0$ efficiently is challenging, as the space of possible missing blocks and their placements is combinatorially large. To guide the search, we leverage the descriptors associated with each stability pattern $R \in \mathcal{R}_{\text{stab}}$: permissible object counts $(n_{\text{base}}, n_{\text{top}})_R$, \textit{supported-by} subgraph $\gG_{\text{supp}}^R$, and subgraphs of geometric relations among the base and top blocks, \ie, $\gG_{\text{base}}^R$ and $\gG_{\text{top}}^R$. Since these descriptors constrain the search space and provide strong priors, we propose a heuristic search guided by stability pattern descriptors to predict missing blocks and relations. To repair an unstable subgraph, we add hidden blocks and relations to either (i) extend the subgraph within the same pattern by preserving descriptor validity, or (ii) evolve it into a more complex pattern that satisfies a richer set of descriptors. The types of added blocks $\typeSet_{\text{hid}}$ are sampled heuristically from existing block types, conditioned on the newly added relations. For example, an unstable {\em \small single-block-stack} may indicate the top block overhangs the base. This can be resolved by converting it to a {\em 
 \small cantilever-with-counterbalance} by changing the relation from {\em \small supported-by-fully} to {\em \small supported-by-partially}, or by transforming it into a {\em \small two-pillar-bridge} by adding a hidden support—of the same type as the base—behind the visible block. In both cases, the guidance of object counts and subgraph constraints from the pattern descriptors enables efficient search.

At each iteration $t$, we first decompose $(\gG_t, \poseSet_t)$ into subgraphs by matching against stability patterns. For each subgraph $(\gG_t^i, \poseSet_t^i)$, we simulate its stability. If unstable, we apply the stability-pattern-guided heuristic search to sample missing blocks and relations. All updates are aggregated into a new abstract relation graph $\gG_{t+1} = \gG_t \cup \gG^{\text{hid}}_t$, which is then re-grounded using compositional diffusion models. This forward-backward refinement continues until a stable structure is found or the maximum number of iterations is reached.

\vspace{-7pt}
\section{Experiment}
\vspace{-7pt}
We evaluate \model\ on $30$ hand-drawn sketches illustrating a wide collection of 3D structures, ranging from iconic buildings (\eg, Marina Bay Sands, Eiffel Tower), to modern house designs (see Appendix~\ref{appendix:test_cases} for all sketches). Generation complexity is quantified by the total number of blocks and the spatial relations in the sketch. We categorize the test cases by whether hidden supports are necessary. \\
We evaluate performance using two metrics: (i) physical stability, measured as the fraction of blocks that remain in place under gravity in simulation; and (ii) resemblance, measured as the fraction of abstract geometric relations from the 2D sketch that are satisfied in the 3D generated structure from its front-view.
Our evaluation tests three hypotheses: \textbf{H1}: Abstract relation graphs enable zero-shot generation of stable, multi-level 3D structures from sketches; \textbf{H2}: Stability-pattern-guided graph updates efficiently recover missing supports; and \textbf{H3}: Sketch-based input provides a more expressive design specification than natural language.

 \begin{figure}[t]
        \centering \hspace{-15pt}\includegraphics[width=0.95\linewidth]{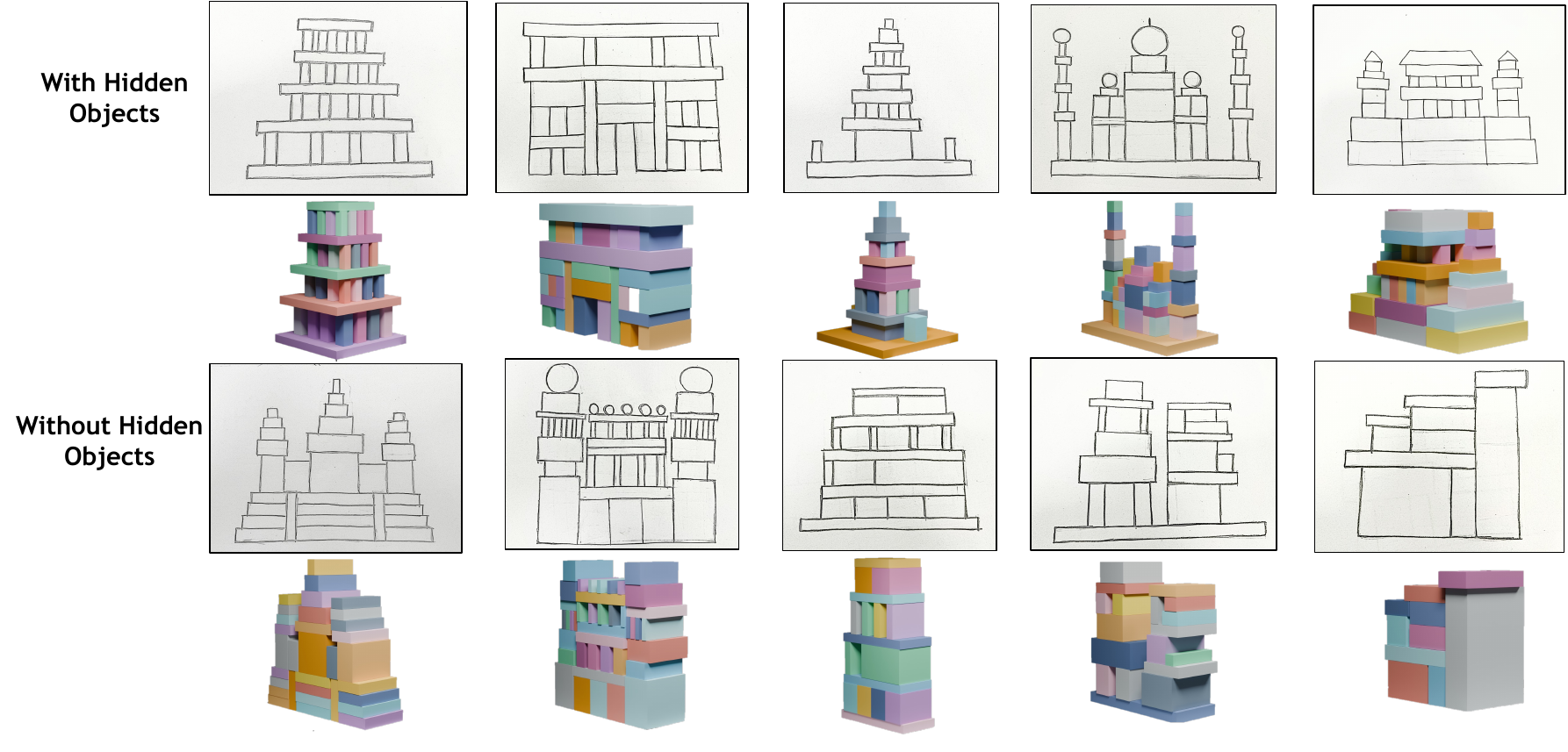}
        \vspace{-7pt}
        \caption[Examples of the 3D Arrangements Generated by StackItUp from 2D Sketches. ]{\textbf{Qualitative results.} 3D arrangements generated by \model from 2D sketches. Top row: sketches where hidden supports are predicted for stability. Bottom row: sketches that require no hidden supports.}
        \vspace{-12pt}
\label{fig:demo_instances}
\end{figure}

\vspace{-7pt}
\subsection{Baselines}
\vspace{-7pt}

We compare \model\ against two baselines and an ablated variant (details in Appendix~\ref{appendix:baseline_implementation}): \\
\shortpara{End-to-End Diffusion Model.}
Following StackGen~\citep{sun2024stackgen}, we implement a single transformer-based diffusion model that directly predicts stable 3D block poses from sketches. The model is trained on the same synthetic dataset as our compositional models, using the extracted 2D sketches as inputs. \\
\shortpara{Direct VLM Prediction.}
Inspired by Blox-net~\citep{goldberg2024blox}, we use a vision-language model (VLM) to predict 3D structures. Given a sketch, we first prompt the VLM to generate a textual description of the scene, then use the description to produce a corresponding 3D block arrangement. \\
\shortpara{Ablation: No Hidden Object Prediction.}
This ablation removes the stability-pattern-guided graph update. It grounds the initial abstract relation graph iteratively using compositional diffusion models when instability is detected, with no addition of hidden supports at each iteration. 

\begin{figure}[t]
    \centering    \includegraphics[width=0.95\linewidth]{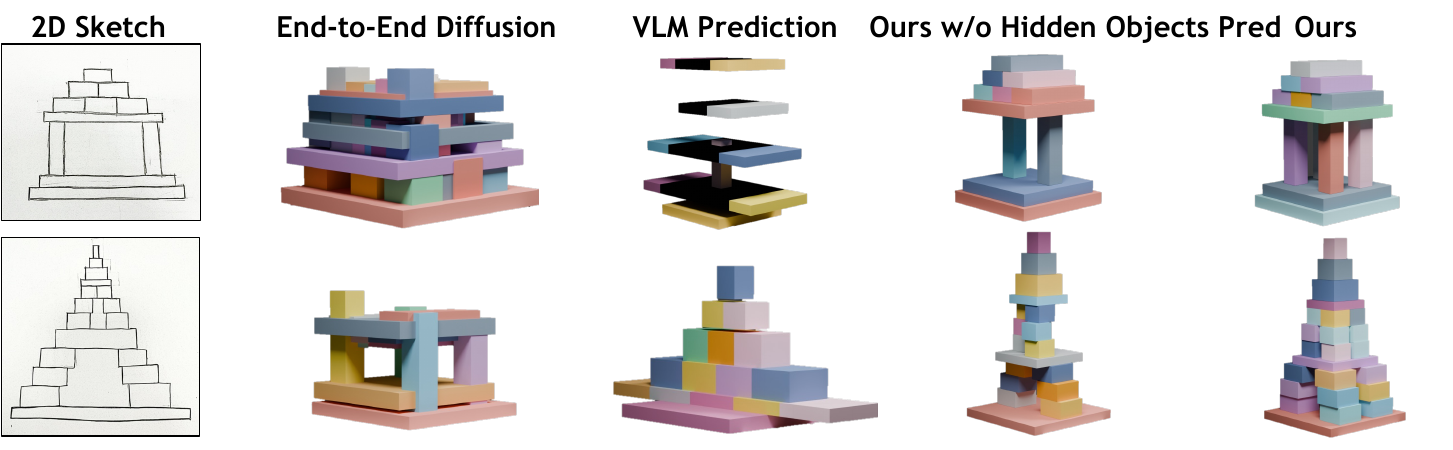}
    \vspace{-5pt}
\caption[StackItUp's Qualitative Comparision with Baselines.]{\textbf{Qualitative Comparison.}Comparison of 3D block arrangements generated from 2D sketches across methods. \model\ consistently produces structures that are both visually faithful and physically stable.}
    \vspace{-15pt}
    \label{fig:qualitative_comp}
\end{figure}

\begin{table*}[tp]
\centering
\caption[Quantitative evaluation of different methods across sketches.]{\textbf{Quantitative evaluation of different methods across sketches.} The reported percentages are the mean and standard deviation over 15 diverse test scenes; the best performance in each column is boldfaced.}
\label{tab:combined_comparison}
\setlength{\tabcolsep}{2pt}
\scriptsize
\resizebox{\textwidth}{!}{
\begin{tabular}{@{}lcccccc@{}}
\toprule
\multirow{2}{*}{\textbf{Model}} & \multicolumn{2}{c}{\textbf{Sketch w/o Hidden Objects}} & \multicolumn{2}{c}{\textbf{Sketch with Hidden Objects}} \\ \cmidrule(lr){2-3} \cmidrule(lr){4-5} 
& Resemblance (\%) & Physical Stability (\%) & Resemblance (\%) & Physical Stability (\%) \\  \midrule

\textbf{End-to-End Diffusion} & 18.2 $_{\pm 7.70}$ & 24.2$_{\pm 13.5}$ & 16.2$_{\pm 9.24}$ & 12.1$_{\pm 8.98}$ \\ 

\textbf{Direct VLM Prediction} & 33.1$_{\pm 22.6}$ & 63.6$_{\pm 26.8}$ & 30.2$_{\pm 26.7}$ & 53.7$_{\pm 35.5}$ \\ \midrule

\textbf{Our Ablation (No Hidden Object Pred)} &  \textbf{84.4$_{\pm 9.45}$} & \textbf{98.8}$_{\pm 4.53}$  & 78.8$_{\pm 17.8}$ &  69.8$_{\pm 33.9}$ \\

\textbf{\model (Ours)} & \textbf{84.4}$_{\pm 9.45}$ & \textbf{98.8}$_{\pm 4.53}$ & \textbf{79.9}$_{\pm 15.0}$ & \textbf{94.5}$_{\pm 15.3}$ \\ \bottomrule
\end{tabular}
}
\vspace{-10pt}
\end{table*}

\vspace{-7pt}
\subsection{Results}
\vspace{-7pt}
\Tabref{tab:combined_comparison} summarizes the overall performance across all test cases. Figures~\ref{fig:demo_instances} and~\ref{fig:qualitative_comp} show example outputs from \model and the baselines. \model\ consistently outperforms all baselines in both physical stability and resemblance to the input sketch, whether or not hidden supports are needed. This validates \textbf{H1}, showing that abstract relation graphs enable effective zero-shot generation of stable, multi-level 3D structures from unseen sketches. In contrast, Direct-VLM-Prediction achieves reasonable stability but poor resemblance, highlighting the importance of precise spatial information in sketches for specifying 3D structures (\textbf{H3}). End-to-end diffusion model performs poorly on both metrics, likely due to the gap between synthetic training data and real-world sketches. Our ablation variant performs comparably to \model\ when hidden supports are not required, but its performance degrades significantly when hidden supports are necessary, underscoring the importance of stability-pattern-guided graph updates and supporting \textbf{H2}. \\
\shortpara{End-to-End Diffusion vs.\ \model: Abstract Relation Graph Enables Zero-Shot Generalization (H1).}
Although both models are trained on the same synthetic dataset, \model\ achieves significantly higher stability and resemblance on test sketches. The test sketches differ substantially from the synthetic training data, causing the end-to-end model to generalize poorly. In contrast, \model\ demonstrates strong zero-shot generalization by leveraging the abstract relation graph as an intermediate representation, which (i) accurately captures the intended geometric relations from arbitrary sketches and (ii) flexibly composes relevant diffusion models at inference to preserve the user's design intent, thus validating \textbf{H1}. \\
\shortpara{Direct-VLM-Prediction vs.\ \model: Sketches Better Capture User Intent (H3).}
While Direct-VLM-Prediction produces stable structures, its low resemblance scores indicate that language descriptions often fail to encode full spatial and geometric intent. In contrast, hand-drawn sketches naturally preserve key geometric and physical relationships that are difficult to express verbally, leading to more faithful reconstructions. This result supports \textbf{H3}. \\
\shortpara{Ablation vs.\ \model: Stability-Pattern-Guided Updates Improve Robustness (H2).}
The ablation removes the stability-pattern-guided graph updates and simply re-optimizes the initial graph using diffusion models without inserting hidden supports. The resulting drop in stability for sketches requiring hidden structures demonstrates that stability-pattern-guided updates provide an efficient and effective heuristic for repairing unstable configurations, supporting \textbf{H2}.

\begin{figure*}[t]
    \centering    \includegraphics[width=0.98\linewidth, page=1]{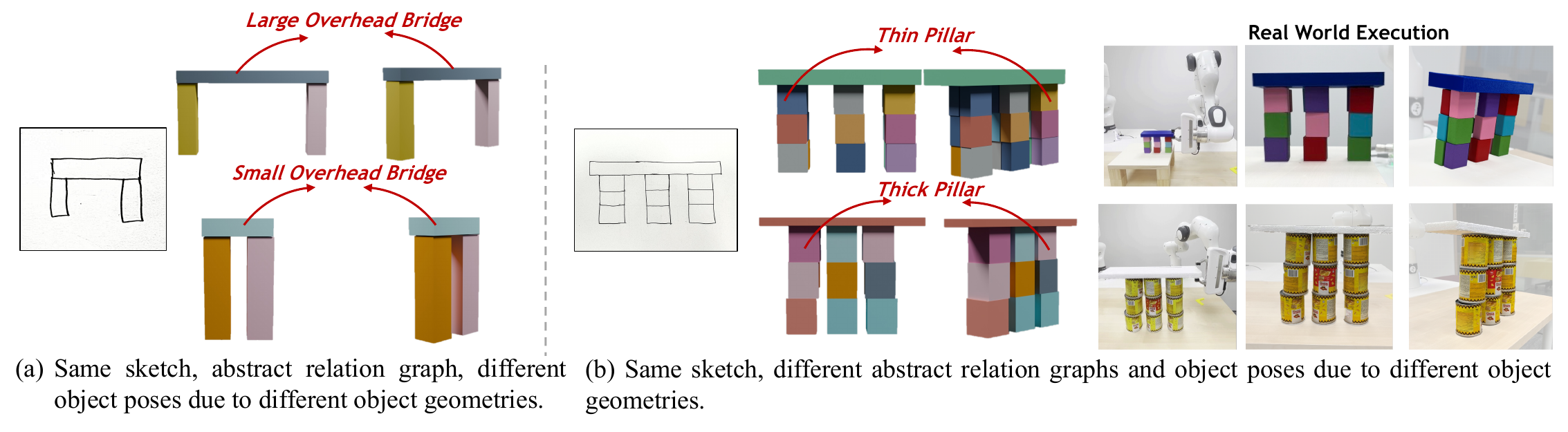}
    \vspace{-5pt}
    \caption[Robustness of StackItUp: adaptation of the 3D pose generation to different block geometries.]{{\small \textbf{Robustness of \model.} \model\ adapts the 3D poses to different block geometries. Given a sketch, it adjusts poses to accommodate geometric variations under the same $\gG$ (a). If the structure is stable through pose adjustment alone, \model\ extends $\gG$ with hidden objects and re-grounds it to new poses (b). }}
    \label{fig:ablation}
    \vspace{-7pt}
\end{figure*}

\begin{figure*}[t!]
\centering\includegraphics[width=\linewidth, page=1]{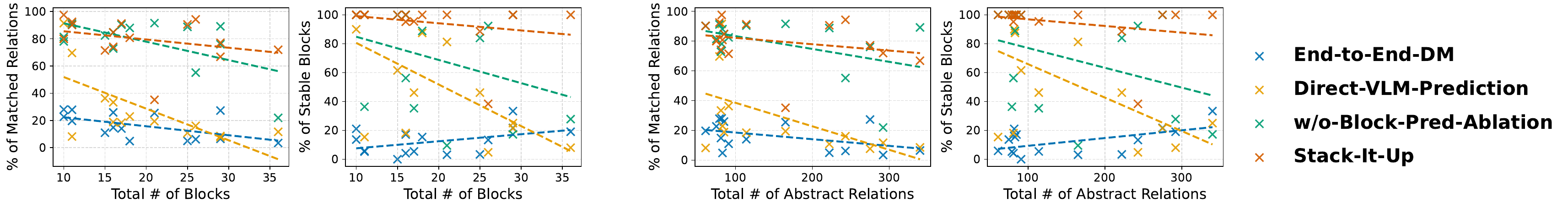}
\vspace{-15pt}
    \caption[Performance \vs\ sketch complexity, measured by the number of blocks and abstract relations.]{Performance \vs\ sketch complexity, measured by the number of blocks and abstract relations.}
    \label{fig:complexity_measure}
\vspace{-15pt}
\end{figure*}
\vspace{-7pt}
\subsection{Showcase of \model's Robustness} 
\vspace{-7pt}
\shortpara{Same Sketch, Different Object Sets: Adapting Object Poses to Satisfy Relations.}
Given the same sketch, \model\ adapts object poses to satisfy the abstract relations even when the candidate block dimensions vary (Figure~\ref{fig:ablation}(a)). This flexibility comes from using the abstract relation graph, which composes diffusion models online to adjust poses according to the specific object geometries. \\
\shortpara{Same Sketch, Different Object Sets: Predicting Support Structures for Stability.}
When the candidate object set changes, \model\ robustly predicts different hidden supports to maintain structural stability (Figure~\ref{fig:ablation}(b)). This is enabled by organizing support structures through stability patterns, which guide the addition of hidden supports when the predicted poses are unstable.\\
\shortpara{Robustness to Sketch Complexity.}
We show how resemblance and stability scores change against the number of blocks and spatial relations in Figure~\ref{fig:complexity_measure}. As complexity increases, Direct-VLM-Prediction degrades sharply, showing the limitations of language-based specification. The end-to-end diffusion model also performs poorly due to the domain gap between real and synthetic sketches. In contrast, \model\ consistently maintains high resemblance and stability, enabled by its abstract relation graph and compositional diffusion models. Unlike the ablation, whose stability declines with larger structures, \model\ effectively predicts hidden supports, demonstrating strong generalization to complex, multi-level designs.

\vspace{-7pt}
\section{Conclusion}
\vspace{-7pt}
\model transforms a rough 2D sketch into a stable 3D block arrangement for robot manipulation. By using an abstract relation graph that captures geometric and stability relations, \model\ enables compositional 3D pose generation and efficient prediction of hidden supports. Experiments on diverse sketches show that \model\ reliably produces stable, visually faithful structures, making sketch-based goal specification accessible to non-expert users.

\vspace{-7pt}
\section{Limitation}
\vspace{-7pt}
While \model\ enables intuitive and robust 3D structure specification from sketches, several limitations point to promising directions for future work.

First, \model\ currently relies on a single 2D front-view sketch, which limits observability of depth and occluded components. While this format is accessible to non-experts, extending the system to support multi-view sketches would improve accuracy. Our abstract relation graph provides a natural abstraction for this extension: integrating sketches from different views reduces to a graph-matching problem, where the key challenge is inferring correspondence between nodes across views. Second, we assume the block types in the sketch are labeled and that candidate object geometries are known. To make \model\ deployable in real-world settings, we could relax these assumptions by integrating 3D perception techniques—such as single-view reconstruction or multi-view fusion—to estimate available object geometries. Then, relative proportions in the sketch can be used to infer likely block types. Finally, closing the loop with real-world robot feedback opens an exciting direction for interactive manipulation. Integrating visual or language-based corrections \citep{Lu2025may,zeng2023learning,sharma2022correcting}, human-in-the-loop refinement \citep{dani2020human,slade2024human,jiang2024llms}, or even learning a reward or cost function \citep{palan2019learning,christiano2017deep,xu2023tidy, MaxEntIRL,DeepMaxEnt,AIRL,xu2022receding,xu2025effective} to discriminate and select among outcomes can make the system more responsive, adaptable, and effective in dynamic environments.



\bibliography{example}  

\begin{thebibliography}{53}
\providecommand{\natexlab}[1]{#1}
\providecommand{\url}[1]{\texttt{#1}}
\expandafter\ifx\csname urlstyle\endcsname\relax
  \providecommand{\doi}[1]{doi: #1}\else
  \providecommand{\doi}{doi: \begingroup \urlstyle{rm}\Url}\fi

\bibitem[Simeon et~al.(2002)Simeon, Cortes, Sahbani, and Laumond]{simeon2002manipulation}
T.~Simeon, J.~Cortes, A.~Sahbani, and J.~Laumond.
\newblock {A Manipulation Planner for Pick and Place Operations Under Continuous Grasps and Placements}.
\newblock In \emph{IEEE Int. Conf. on Robotics \& Automation}, 2002.

\bibitem[Holladay et~al.(2013)Holladay, Barry, Kaelbling, and Lozano-Pérez]{6631099}
A.~Holladay, J.~Barry, L.~P. Kaelbling, and T.~Lozano-Pérez.
\newblock {Object Placement as Inverse Motion Planning}.
\newblock In \emph{IEEE Int. Conf. on Robotics \& Automation}, 2013.

\bibitem[Suárez-Ruiz and Pham(2016)]{suarez-ruiz2016framework}
F.~Suárez-Ruiz and Q.-C. Pham.
\newblock {A Framework for Fine Robotic Assembly}.
\newblock In \emph{IEEE Int. Conf. on Robotics \& Automation}, 2016.

\bibitem[Zucker et~al.(2013)Zucker, Ratliff, Dragan, Pivtoraiko, Klingensmith, Dellin, Bagnell, and Srinivasa]{zucker2013chomp}
M.~Zucker, N.~Ratliff, A.~D. Dragan, M.~Pivtoraiko, M.~Klingensmith, C.~M. Dellin, J.~A. Bagnell, and S.~S. Srinivasa.
\newblock {CHOMP: Covariant Hamiltonian Optimization for Motion Planning}.
\newblock \emph{Int. J. Robotics Research}, 32, 2013.

\bibitem[Karaman and Frazzoli(2011)]{karaman2011sampling}
S.~Karaman and E.~Frazzoli.
\newblock {Sampling-Based Algorithms for Optimal Motion Planning}.
\newblock \emph{Int. J. Robotics Research}, 30, 2011.

\bibitem[Kavraki et~al.(1996)Kavraki, Svestka, Latombe, and Overmars]{kavraki1996probabilistic}
L.~E. Kavraki, P.~Svestka, J.-C. Latombe, and M.~H. Overmars.
\newblock {Probabilistic Roadmaps for Path Planning in High-Dimensional Configuration Spaces}.
\newblock \emph{IEEE Trans. on Robotics and Automation}, 12, 1996.

\bibitem[LaValle(1998)]{lavalle1998rapidly}
S.~LaValle.
\newblock {Rapidly-Exploring Random Trees: A New Tool for Path Planning}.
\newblock \emph{Research Report 9811}, 1998.

\bibitem[Babi{\v{c}} and Rebolj(2016)]{babivc2016culture}
N.~{\v{C}}. Babi{\v{c}} and D.~Rebolj.
\newblock {Culture Change in Construction Industry: From 2D Toward BIM Based Construction}.
\newblock \emph{Journal of Information Technology in Construction (ITcon)}, 21, 2016.

\bibitem[Hunt(2013)]{hunt2013bim}
C.~A. Hunt.
\newblock {The Benefits of Using Building Information Modeling in Structural Engineering}.
\newblock \url{https://digitalcommons.usu.edu/gradreports/319}, 2013.

\bibitem[{Novel BIM}(2023)]{novelbim2023problems}
{Novel BIM}.
\newblock {Problems of 2D Drawings in Construction Projects}, 2023.

\bibitem[{Mars BIM}(2023)]{marsbim2023drawings}
{Mars BIM}.
\newblock {Traditional Drawings Vs BIM Technology: Why 2D Plans Are No Longer Enough}, 2023.

\bibitem[{Tekla}(2023)]{tekla2023harming}
{Tekla}.
\newblock {Are You Harming Your Business by Sticking with 2D?}, 2023.

\bibitem[{Shalin Designs}(2023)]{shalin2023modeling}
{Shalin Designs}.
\newblock {Importance of Structural 3D Modeling}, 2023.

\bibitem[Sen et~al.(2023)Sen, Agarwal, Singh, B., Sridhar, and Krishna]{10160365}
B.~Sen, A.~Agarwal, G.~Singh, B.~B., S.~Sridhar, and M.~Krishna.
\newblock {SCARP: 3D Shape Completion in Arbitrary Poses for Improved Grasping}.
\newblock In \emph{IEEE Int. Conf. on Robotics \& Automation}, 2023.

\bibitem[Irshad et~al.(2022)Irshad, Zakharov, Ambrus, Kollar, Kira, and Gaidon]{irshad2022shapo}
M.~Z. Irshad, S.~Zakharov, R.~Ambrus, T.~Kollar, Z.~Kira, and A.~Gaidon.
\newblock {ShAPO: Implicit Representations for Multi-Object Shape, Appearance, and Pose Optimization}.
\newblock In \emph{Proc. European Conference on Computer Vision}, 2022.

\bibitem[Garrett et~al.(2018{\natexlab{a}})Garrett, Lozano-Perez, and Kaelbling]{garrett2018ffrob}
C.~R. Garrett, T.~Lozano-Perez, and L.~P. Kaelbling.
\newblock {FFRob: Leveraging Symbolic Planning for Efficient Task and Motion Planning}.
\newblock \emph{Int. J. Robotics Research}, 37, 2018{\natexlab{a}}.

\bibitem[Garrett et~al.(2018{\natexlab{b}})Garrett, Lozano-P{\'e}rez, and Kaelbling]{garrett2018sampling}
C.~R. Garrett, T.~Lozano-P{\'e}rez, and L.~P. Kaelbling.
\newblock {Sampling-Based Methods for Factored Task and Motion Planning}.
\newblock \emph{Int. J. Robotics Research}, 37, 2018{\natexlab{b}}.

\bibitem[Yang et~al.(2023)Yang, Mao, Du, Wu, Tenenbaum, Lozano-P{\'e}rez, and Kaelbling]{yang2023diffusion}
Z.~Yang, J.~Mao, Y.~Du, J.~Wu, J.~B. Tenenbaum, T.~Lozano-P{\'e}rez, and L.~P. Kaelbling.
\newblock {Compositional Diffusion-Based Continuous Constraint Solvers}.
\newblock In \emph{Conference on Robot Learning}, 2023.

\bibitem[Xu et~al.(2024)Xu, Mao, Du, Lozano-Pérez, Kaelbling, and Hsu]{Xu-RSS-24}
Y.~Xu, J.~Mao, Y.~Du, T.~Lozano-Pérez, L.~P. Kaelbling, and D.~Hsu.
\newblock {Set It Up!: Functional Object Arrangement with Compositional Generative Models}.
\newblock In \emph{Proc. Robotics: Science \& Systems}, 2024.

\bibitem[Zhu et~al.(2021)Zhu, Tremblay, Birchfield, and Zhu]{zhu2021hierarchical}
Y.~Zhu, J.~Tremblay, S.~Birchfield, and Y.~Zhu.
\newblock {Hierarchical Planning for Long-Horizon Manipulation with Geometric and Symbolic Scene Graphs}.
\newblock In \emph{IEEE Int. Conf. on Robotics \& Automation}, 2021.

\bibitem[Du et~al.(2023)Du, Durkan, Strudel, Tenenbaum, Dieleman, Fergus, Sohl-Dickstein, Doucet, and Grathwohl]{du2023reduce}
Y.~Du, C.~Durkan, R.~Strudel, J.~B. Tenenbaum, S.~Dieleman, R.~Fergus, J.~Sohl-Dickstein, A.~Doucet, and W.~Grathwohl.
\newblock {Reduce, Reuse, Recycle: Compositional Generation with Energy-Based Diffusion Models and MCMC}.
\newblock In \emph{Proc. Int. Conf. on Machine Learning}, 2023.

\bibitem[Sj{\"o}berg et~al.(2023)Sj{\"o}berg, Lindqvist, {\"O}nnheim, Jirstrand, and Svensson]{sjoberg2023mcmc}
A.~Sj{\"o}berg, J.~Lindqvist, M.~{\"O}nnheim, M.~Jirstrand, and L.~Svensson.
\newblock {MCMC-Correction of Score-Based Diffusion Models for Model Composition}.
\newblock \emph{arXiv preprint arXiv:2307.14012}, 2023.

\bibitem[Barber et~al.(2010)Barber, Shucksmith, MacDonald, and W{\"u}nsche]{barber2010sketch}
C.~M. Barber, R.~J. Shucksmith, B.~MacDonald, and B.~C. W{\"u}nsche.
\newblock {Sketch-Based Robot Programming}.
\newblock In \emph{IEEE Int. Conf. on Image and Vision Computing New Zealand}, 2010.

\bibitem[Porfirio et~al.(2023)Porfirio, Stegner, Cakmak, Saupp{\'e}, Albarghouthi, and Mutlu]{porfirio2023sketching}
D.~Porfirio, L.~Stegner, M.~Cakmak, A.~Saupp{\'e}, A.~Albarghouthi, and B.~Mutlu.
\newblock {Sketching Robot Programs on the Fly}.
\newblock In \emph{ACM/IEEE Int. Conf. on Human-Robot Interaction}, 2023.

\bibitem[Sundaresan et~al.(2024)Sundaresan, Vuong, Gu, Xu, Xiao, Kirmani, Yu, Stark, Jain, Hausman, et~al.]{sundaresan2024rt}
P.~Sundaresan, Q.~Vuong, J.~Gu, P.~Xu, T.~Xiao, S.~Kirmani, T.~Yu, M.~Stark, A.~Jain, K.~Hausman, et~al.
\newblock {RT-Sketch: Goal-Conditioned Imitation Learning From Hand-Drawn Sketches}.
\newblock In \emph{Conference on Robot Learning}, 2024.

\bibitem[Cui et~al.(2022)Cui, Niekum, Gupta, Kumar, and Rajeswaran]{cui2022can}
Y.~Cui, S.~Niekum, A.~Gupta, V.~Kumar, and A.~Rajeswaran.
\newblock {Can Foundation Models Perform Zero-Shot Task Specification for Robot Manipulation?}
\newblock In \emph{Learning for Dynamics and Control Conference}, 2022.

\bibitem[Gu et~al.(2023)Gu, Kirmani, Wohlhart, Lu, Arenas, Rao, Yu, Fu, Gopalakrishnan, Xu, et~al.]{gu2023robotic}
J.~Gu, S.~Kirmani, P.~Wohlhart, Y.~Lu, M.~G. Arenas, K.~Rao, W.~Yu, C.~Fu, K.~Gopalakrishnan, Z.~Xu, et~al.
\newblock {Robotic Task Generalization Via Hindsight Trajectory Sketches}.
\newblock In \emph{First Workshop on Out-of-Distribution Generalization in Robotics at CoRL 2023}, 2023.

\bibitem[Goldberg et~al.(2024)Goldberg, Kondap, Qiu, Ma, Fu, Kerr, Huang, Chen, Fang, and Goldberg]{goldberg2024blox}
A.~Goldberg, K.~Kondap, T.~Qiu, Z.~Ma, L.~Fu, J.~Kerr, H.~Huang, K.~Chen, K.~Fang, and K.~Goldberg.
\newblock {Blox-Net: Generative Design-for-Robot-Assembly Using VLM Supervision, Physics Simulation, and a Robot with Reset}.
\newblock \emph{arXiv preprint arXiv:2409.17126}, 2024.

\bibitem[Badagabettu et~al.(2024)Badagabettu, Yarlagadda, and Farimani]{badagabettu2024query2cad}
A.~Badagabettu, S.~S. Yarlagadda, and A.~B. Farimani.
\newblock {Query2CAD: Generating CAD Models Using Natural Language Queries}.
\newblock \emph{arXiv preprint arXiv:2406.00144}, 2024.

\bibitem[Hong et~al.(2024)Hong, Zhang, Gu, Bi, Zhou, Liu, Liu, Sunkavalli, Bui, and Tan]{hong2024lrm}
Y.~Hong, K.~Zhang, J.~Gu, S.~Bi, Y.~Zhou, D.~Liu, F.~Liu, K.~Sunkavalli, T.~Bui, and H.~Tan.
\newblock {LRM: Large Reconstruction Model for Single Image to 3D}.
\newblock In \emph{Int. Conf. on Learning Representations}, 2024.

\bibitem[Tang et~al.(2024)Tang, Ren, Zhou, Liu, and Zeng]{tang2024dreamgaussian}
J.~Tang, J.~Ren, H.~Zhou, Z.~Liu, and G.~Zeng.
\newblock {DreamGaussian: Generative Gaussian Splatting for Efficient 3D Content Creation}.
\newblock In \emph{Int. Conf. on Learning Representations}, 2024.

\bibitem[Sun et~al.(2024)Sun, Yoneda, Wheeler, Jiang, and Walter]{sun2024stackgen}
L.~Sun, T.~Yoneda, S.~W. Wheeler, T.~Jiang, and M.~R. Walter.
\newblock {Stackgen: Generating Stable Structures From Silhouettes Via Diffusion}.
\newblock \emph{arXiv preprint arXiv:2409.18098}, 2024.

\bibitem[Liu et~al.(2024)Liu, Lin, Liu, Long, Dou, Guo, Luo, and Wang]{liu2024part123}
A.~Liu, C.~Lin, Y.~Liu, X.~Long, Z.~Dou, H.-X. Guo, P.~Luo, and W.~Wang.
\newblock {Part123: Part-Aware 3D Reconstruction From a Single-View Image}.
\newblock In \emph{ACM SIGGRAPH Conference Papers}, 2024.

\bibitem[Liu et~al.(2023{\natexlab{a}})Liu, Wu, Van~Hoorick, Tokmakov, Zakharov, and Vondrick]{liu2023zero}
R.~Liu, R.~Wu, B.~Van~Hoorick, P.~Tokmakov, S.~Zakharov, and C.~Vondrick.
\newblock {Zero-1-to-3: Zero-Shot One Image to 3D Object}.
\newblock In \emph{IEEE Int. Conf. on Computer Vision}, 2023{\natexlab{a}}.

\bibitem[Liu et~al.(2023{\natexlab{b}})Liu, Xu, Jin, Chen, Varma~T, Xu, and Su]{liu2023one}
M.~Liu, C.~Xu, H.~Jin, L.~Chen, M.~Varma~T, Z.~Xu, and H.~Su.
\newblock {One-2-3-45: Any Single Image to 3D Mesh in 45 Seconds Without Per-Shape Optimization}.
\newblock \emph{Advances in Neural Information Processing Systems}, 2023{\natexlab{b}}.

\bibitem[Liu et~al.(2024)Liu, Shi, Chen, Zhang, Xu, Wei, Chen, Zeng, Gu, and Su]{liu2024one}
M.~Liu, R.~Shi, L.~Chen, Z.~Zhang, C.~Xu, X.~Wei, H.~Chen, C.~Zeng, J.~Gu, and H.~Su.
\newblock {One-2-3-45++: Fast Single Image to 3D Objects with Consistent Multi-View Generation and 3D Diffusion}.
\newblock In \emph{IEEE Conf. on Computer Vision \& Pattern Recognition}, 2024.

\bibitem[Ho et~al.(2020)Ho, Jain, and Abbeel]{ho2020denoising}
J.~Ho, A.~Jain, and P.~Abbeel.
\newblock {Denoising Diffusion Probabilistic Models}.
\newblock In \emph{Advances in Neural Information Processing Systems}, 2020.

\bibitem[Lu et~al.(2025)Lu, Ma, Hori, and Romeres]{Lu2025may}
K.~Lu, C.~Ma, C.~Hori, and D.~Romeres.
\newblock {KitchenVLA: Iterative Vision‑Language Corrections for Robotic Execution of Human Tasks}.
\newblock In \emph{IEEE International Conference on Robotics and Automation Workshop on Safely Leveraging Vision‑Language Foundation Models in Robotics (SafeLVMs@ICRA)}, May 2025.
\newblock URL \url{https://www.merl.com/publications/TR2025-068}.

\bibitem[Zeng and Xu(2023)]{zeng2023learning}
Y.~Zeng and Y.~Xu.
\newblock Learning reward for physical skills using large language model.
\newblock \emph{arXiv preprint arXiv:2310.14092}, 2023.

\bibitem[Sharma et~al.(2022)Sharma, Sundaralingam, Blukis, Paxton, Hermans, Torralba, Andreas, and Fox]{sharma2022correcting}
P.~Sharma, B.~Sundaralingam, V.~Blukis, C.~Paxton, T.~Hermans, A.~Torralba, J.~Andreas, and D.~Fox.
\newblock Correcting robot plans with natural language feedback.
\newblock \emph{arXiv preprint arXiv:2204.05186}, 2022.

\bibitem[Dani et~al.(2020)Dani, Salehi, Rotithor, Trombetta, and Ravichandar]{dani2020human}
A.~P. Dani, I.~Salehi, G.~Rotithor, D.~Trombetta, and H.~Ravichandar.
\newblock Human-in-the-loop robot control for human-robot collaboration: Human intention estimation and safe trajectory tracking control for collaborative tasks.
\newblock \emph{IEEE Control Systems Magazine}, 40\penalty0 (6):\penalty0 29--56, 2020.

\bibitem[Slade et~al.(2024)Slade, Atkeson, Donelan, Houdijk, Ingraham, Kim, Kong, Poggensee, Riener, Steinert, et~al.]{slade2024human}
P.~Slade, C.~Atkeson, J.~M. Donelan, H.~Houdijk, K.~A. Ingraham, M.~Kim, K.~Kong, K.~L. Poggensee, R.~Riener, M.~Steinert, et~al.
\newblock On human-in-the-loop optimization of human--robot interaction.
\newblock \emph{Nature}, 633\penalty0 (8031):\penalty0 779--788, 2024.

\bibitem[Jiang et~al.(2024)Jiang, Xu, and Hsu]{jiang2024llms}
C.~Jiang, Y.~Xu, and D.~Hsu.
\newblock Llms for robotic object disambiguation.
\newblock \emph{arXiv preprint arXiv:2401.03388}, 2024.

\bibitem[Palan et~al.(2019)Palan, Landolfi, Shevchuk, and Sadigh]{palan2019learning}
M.~Palan, N.~C. Landolfi, G.~Shevchuk, and D.~Sadigh.
\newblock Learning reward functions by integrating human demonstrations and preferences.
\newblock \emph{arXiv preprint arXiv:1906.08928}, 2019.

\bibitem[Christiano et~al.(2017)Christiano, Leike, Brown, Martic, Legg, and Amodei]{christiano2017deep}
P.~F. Christiano, J.~Leike, T.~Brown, M.~Martic, S.~Legg, and D.~Amodei.
\newblock Deep reinforcement learning from human preferences.
\newblock In \emph{Advances in Neural Information Processing Systems}, 2017.

\bibitem[Xu and Hsu(2023)]{xu2023tidy}
Y.~Xu and D.~Hsu.
\newblock How to tidy up a table: Fusing visual and semantic commonsense reasoning for robotic tasks with vague objectives.
\newblock \emph{arXiv preprint arXiv:2307.11319}, 2023.

\bibitem[Ziebart et~al.(2008)Ziebart, Maas, Bagnell, and Dey]{MaxEntIRL}
B.~D. Ziebart, A.~Maas, J.~A. Bagnell, and A.~K. Dey.
\newblock Maximum entropy inverse reinforcement learning.
\newblock 2008.

\bibitem[Wulfmeier et~al.(2016)Wulfmeier, Ondruska, and Posner]{DeepMaxEnt}
M.~Wulfmeier, P.~Ondruska, and I.~Posner.
\newblock Maximum entropy deep inverse reinforcement learning.
\newblock In \emph{Advances in Neural Information Processing Systems}, 2016.

\bibitem[Fu et~al.(2018)Fu, Luo, and Levine]{AIRL}
J.~Fu, K.~Luo, and S.~Levine.
\newblock Learning robust rewards with adversarial inverse reinforcement learning.
\newblock In \emph{Int. Conf. on Learning Representations}, 2018.

\bibitem[Xu et~al.(2022)Xu, Gao, and Hsu]{xu2022receding}
Y.~Xu, W.~Gao, and D.~Hsu.
\newblock Receding horizon inverse reinforcement learning.
\newblock In \emph{Advances in Neural Information Processing Systems}, 2022.

\bibitem[Xu et~al.(2025)Xu, Doshi-Velez, and Hsu]{xu2025effective}
Y.~Xu, F.~Doshi-Velez, and D.~Hsu.
\newblock On the effective horizon of inverse reinforcement learning.
\newblock In \emph{Proceedings of the 24th International Conference on Autonomous Agents and Multiagent Systems}, pages 2208--2216, 2025.

\bibitem[Vincent(2011)]{vincent2011connection}
P.~Vincent.
\newblock {A Connection Between Score Matching and Denoising Autoencoders}.
\newblock \emph{Neural Comput.}, 23\penalty0 (7):\penalty0 1661--1674, 2011.

\bibitem[Coleman et~al.(2014)Coleman, Sucan, Chitta, and Correll]{coleman2014reducing}
D.~Coleman, I.~Sucan, S.~Chitta, and N.~Correll.
\newblock Reducing the barrier to entry of complex robotic software: a {MoveIt!} case study.
\newblock \emph{Journal of Software Engineering for Robotics}, 2014.

\end{thebibliography}
\clearpage
\appendix
\section{Abstract Relation Library}
\label{appendix:abstract_relation_library}

The abstract relation library comprises a suite of qualitative geometric relations and local stability patterns used to compactly represent multi-layer block structures. Each relation is defined as a classifier $h_R$ that examines block dimensions and relative poses to determine if a specific relation holds. This systematic encoding enables straightforward extraction of relation graphs from block arrangements and provides a foundation for generating synthetic data and training the class-conditional diffusion models $f_R$. Table~\ref{tab:relationships} lists the symbolic names and arities of all relations. Below, we detail the implementation of each classifier, following the rules that also underlie our synthetic data generation.

\subsection{Geometric Relations}\label{appendix:geometric_relation}

We define 24 qualitative geometric relations in total. Of these, 13 are front-view ($x$–$z$ plane) relations, such as \emph{left-of} and \emph{horizontally-aligned}, which can be computed directly from the bounding boxes in the sketch. The remaining 11 describe layout in the $x$–$y$ (depth) plane (e.g., \emph{front-of}, \emph{depth-aligned}), and are only predicted once hidden (occluded) supports are inferred. For each relation, we specify its arity, associated plane, language description, and the explicit rule used by its classifier $h_R$. 

\begin{longtable}[ht]{>{\raggedright\arraybackslash}p{2cm}>{\raggedright\arraybackslash}p{1cm}>
{\raggedright\arraybackslash}p{1cm}>{\raggedright\arraybackslash}p{4cm}>{\raggedright\arraybackslash}p{4cm}}
\toprule
\textbf{Relation} & \textbf{Arity} & \textbf{Plane} & \textbf{Description} & \textbf{Implementation of $h_R$} \\
\midrule
left-of($o_A$, $o_B$) & 2 & $x$-$z$ & 
Block $o_A$ is positioned entirely to the left of block $o_B$ on the same support level; they do not overlap along the $x$-axis, but their front-to-back positions overlap substantially, meaning they are laterally offset and adjacent as seen from the front. &
$o_A$.right\_x $\leq$ $o_B$.left\_x $+$ EPS; $|$right\_x $-$ left\_x$| <$ GAP; $y$-projections overlap $>$ ALPHA $\times$ min depth; same $z$-level\\

\midrule
left-in($o$, table) & 2 & $x$-$z$ &
Block $o$ is located entirely to the left side of the table, with its rightmost side not crossing the table center. &
$o$.right\_x $<$ table.center\_x \\

\midrule
right-in($o$, table) & 2 & $x$-$z$ &
Block $o$ is located entirely to the right side of the table, with its leftmost side not crossing the table center. &
$o$.left\_x $>$ table.center\_x \\

\midrule
center-in($o$, table) & 2 & $x$-$z$ &
Block $o$ is centered on the table, having its center positioned at or very close to the table's origin. &
$|o$.center\_x-table.center\_x$| <$ EPS and $|o$.center\_y-table.center\_y$| <$ EPS \\

\midrule

supported-by-partially($o_A$, $o_B$) & 2 & $x$-$z$ &
Block $o_A$ sits on top of (is immediately above) block $o_B$ but its base is only partially resting on $o_B$'s top surface—so some but not all of its footprint is supported. &
$o_A$.base\_z $\approx$ $o_B$.top\_z (within EPS), $o_A$'s $xy$ footprint overlaps with but is not contained in $o_B$'s \\

\midrule
supported-by-fully($o_A$, $o_B$) & 2 & $x$-$z$ &
Block $o_A$ sits on top of (is immediately above) block $o_B$, and its entire base lies within (is fully supported by) $o_B$'s top surface. &
$o_A$.base\_z $\approx$ $o_B$.top\_z (within EPS), $o_A$'s $xy$ footprint is fully contained in $o_B$'s \\

\midrule
horizontal-aligned($o_A$, $o_B$) & 2 & $x$-$z$ &
Blocks $o_A$ and $o_B$ have the same front-back ($y$) coordinate—either at their centers or at matching front/back surfaces—indicating that they’re horizontally aligned across the table (as seen from the front). &
$|$center\_y$_A - $center\_y$_B| <$ EPS; or front/back surfaces match \\

\midrule
vertical-aligned-centroid($o_A$, $o_B$) & 2 & $x$-$z$ &
Block $o_A$ is stacked directly above block $o_B$; both their $x$ and $y$ centroids align, so they form a straight vertical column. &
$o_A$.base\_z $\approx$ $o_B$.top\_z; 
$|(x_A,y_A)-(x_B,y_B)|<$ EPS\\

\midrule
vertical-aligned-left($o_A$, $o_B$) & 2 & $x$-$z$ &
Blocks $o_A$ and $o_B$ are precisely stacked so that their left ($x$) sides line up, and they overlap significantly along the $y$ axis (as seen from above). &
$o_A$.base\_z $\approx$ $o_B$.top\_z; $|$left\_x$_A - $left\_x$_B|<$EPS; $y$-overlap $>$ALPHA$\times$depth\\

\midrule
vertical-aligned-right($o_A$, $o_B$) & 2 & $x$-$z$ &
Blocks $o_A$ and $o_B$ are stacked so their right ($x$) sides line up exactly, with significant $y$-overlap (depth). &
$o_A$.base\_z $\approx$ $o_B$.top\_z; $|$right\_x$_A - $right\_x$_B|<$EPS; $y$-overlap $>$ALPHA$\times$depth\\

\midrule
horizontal-aligned-in-a-line($o_1$, …, $o_n$) & $n$ & $x$-$z$ &
Several blocks are arranged in a perfectly straight row (line) horizontally (left to right) with exactly matched $y$ positions and equal $z$; typical for bridges/beams. &
All $|$center\_y$_i - $center\_y$_j| <$ EPS; $x$ positions ordered, spacing regular \\
\midrule
touching-along-x($o_A$, $o_B$) & 2 & $x$-$z$ & 
Blocks $o_A$ and $o_B$ are side by side and touch exactly at their adjoining left/right faces, with a strong front-back ($y$) overlap. &
$|$right\_x$_A - $left\_x$_B| <$EPS, $y$-overlap $>$ALPHA$\times$min depth\\

\midrule
near-along-x($o_A$, $o_B$) & 2 & $x$-$z$ & 
Blocks $o_A$ and $o_B$ are on the same level and are placed close to each other side-by-side along the left-right ($x$) direction, but with a small gap (not touching). They overlap substantially in the front-back ($y$) direction, meaning they are nearly “neighbors” from the front view but not actually in contact with each other. &
EPS $\leq$ gap\_x $<$ D\_NEAR; $y$-projection overlap $>$ ALPHA $\times$ min depth; same $z$-level \\

\midrule
front-of($o_A$, $o_B$) & 2 & $x$-$y$ &
Block $o_A$ is positioned entirely in front of block $o_B$ on the same level; their sides overlap along left/right, but $o_A$ is closer to the front (observer), not overlapping with $o_B$ in the $y$ direction. &
$o_A$.back\_y $\leq$ $o_B$.front\_y $+$ EPS; $|$back\_y $-$ front\_y$| <$ GAP; $x$-projections overlap $>$ BETA $\times$ min width; same $z$-level \\

\midrule
front-in($o$, table) & 2 & $x$-$y$ & 
Block $o$ is positioned entirely in front of the table's center (the $y=0$ axis), meaning its entire back face is still in front of the table origin. The block lies between the observer and the center of the table, not straddling or exceeding the central axis. &
$o$.back\_y $<$ table.center\_y \\

\midrule
back-in($o$, table) & 2 & $x$-$y$ &
Block $o$ is located entirely behind the table's origin, with its frontmost point behind the $y=0$ axis. &
$o$.front\_y $>$ table.center\_y \\

\midrule

touching-along-y($o_A$, $o_B$) & 2 & $x$-$y$ & 
Blocks $o_A$ and $o_B$ are placed side by side in the front-back ($y$) axis, so that one’s back directly meets the other’s front, with strong overlap along the left-right ($x$) axis; they “touch” along their $y$ faces. &
$|$back\_y $-$ front\_y$| <$ EPS; $x$-projection overlap $>$ ALPHA$\times$min width \\
\midrule
near-along-y($o_A$, $o_B$) & 2 & $x$-$y$ &
Blocks $o_A$ and $o_B$ are positioned nearly touching in the front-back ($y$) direction—separated by a small gap, but otherwise closely aligned, and substantially overlapping along the $x$ axis. &
EPS $\leq$ gap\_y $<$ D\_NEAR; $x$-projection overlap $>$ ALPHA$\times$min width \\

\midrule
depth-aligned($o_A$, $o_B$) & 2 & $x$-$y$ &
Blocks $o_A$ and $o_B$ have aligned depth (front-back) placements: their centers, or edges, in the left-right ($x$) axis coincide; often used for checking columnar or grid-like arrangements. &
$|$center\_x$_A - $center\_x$_B| <$ EPS; or for left/right versions, $|$left/right\_x$_A - $left/right\_x$_B| <$ EPS\\

\midrule
depth-aligned-in-a-line($o_1$, …, $o_n$) & $n$ & $x$-$y$ &
A group of $n$ blocks, arranged in a straight line along the $y$ (front-back) axis, each with the same $x$ position (column formation), often with similar or equal spacing between centers. &
All $|$center\_x$_i - $center\_x$_j| <$ EPS; $y$ positions ordered, spacing regular\\

\midrule
regular-grid-sparse($o_1$, …, $o_n$) & $n$ & $x$-$y$ &
A set of blocks forms a 2D grid in the $x$-$y$ plane, where the left-right and front-back spacings between blocks are consistent but relatively wide—leaving space between adjacent blocks. &
All blocks similar size, grouped into rows/columns by $x/y$; adjacent grid pairs are more than just touching (spacing$>$touch\_eps), but rows/cols are regular\\

\midrule
regular-grid-compact($o_1$, …, $o_n$) & $n$ & $x$-$y$ &
A set of blocks arranged in a closely packed, regular 2D grid, so that each block touches its neighbors both horizontally and vertically without gaps, and fills nearly all of the bounding rectangle. &
All blocks similar size, assigned to rows/columns by $x/y$; all adjacent blocks touch (spacing $\leq$ touch\_eps); grid fills $\geq$95\% of bounding box\\

\midrule
random-split-grid-sparse($o_1$, …, $o_n$) & $n$ & $x$-$y$ &
A group of blocks covers much—but not all—of a region in the $x$-$y$ plane, forming a loosely connected configuration that is not fully regular, but no large gaps exist; may result from a random split/composition. &
Sum of all block areas $\geq$90\% of total bounding box; block positions irregular\\

\midrule
random-split-grid-compact($o_1$, …, $o_n$) & $n$ & $x$-$y$ &
A group of blocks forms a compact area in the $x$-$y$ plane, filling almost all available space but without the strict regularity of a grid. &
Sum of all block areas $\geq$ 90\% of bounding box; positions irregular, but very little wasted space\\

\bottomrule
\label{tab:geometric_relation_dictionary}
\end{longtable}

\subsection{Stability Patterns}\label{appendix:stability_pattern}

\begin{figure*}[hp]
    \centering
    \includegraphics[width=\linewidth, page=1]{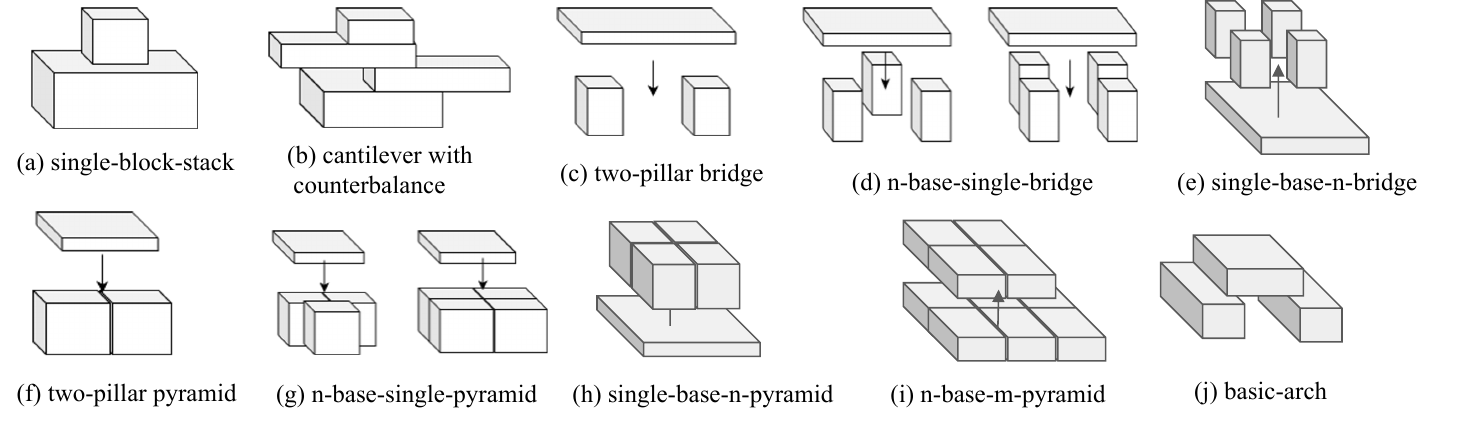}
    \caption[Illustration of the ten stability patterns.]{Illustration of the ten stability patterns.}
    \label{fig:stability_patterns}
\end{figure*}

We define 10 local stability patterns, each describing a recurring two-tier stable arrangement (e.g., pillar-bridge, stack). Each pattern’s classifier $h_R$ is characterized by four descriptors: (i) allowed counts of base and top blocks $({n_{\text{base}}}, {n_{\text{top}}})$; (ii) admissible supported-by subgraphs ${\gG_{\text{supp}}^R}$ connecting base and top blocks; (iii) required or allowed interrelations among the base blocks ${\gG_{\text{base}}^R}$; and (iv) likewise among top blocks ${\gG_{\text{top}}^R}$. These descriptors allow the automatic extraction of stability patterns from block arrangements, support straightforward synthetic data generation, and guide the compositional assembly of stable multi-level structures. By enumerating these patterns and their descriptors, we enable principled graph growth and hidden block insertion subject to stability constraints. Detailed specification for each stability classifier appears below.

\begin{longtable}[ht]{>{\raggedright\arraybackslash}p{2.5cm}>{\raggedright\arraybackslash}p{3.3cm}>{\raggedright\arraybackslash}p{1.7cm}>{\raggedright\arraybackslash}p{4cm}}
\toprule
\textbf{Stability Pattern} & \textbf{Natural Language Description} & \textbf{Block Counts (base, top)} & \textbf{Subgraph Pattern Constraints} \\
\midrule
\multicolumn{4}{p{\dimexpr\linewidth-2\tabcolsep\relax}}{
\textbf{Notes on Subgraph Patterns:} 
\par
\textbf{(a) Supported-by subgraph}: Describes which “supported-by-fully” and “supported-by-partially” relations between base and top blocks must hold.
\par
\textbf{(b) Base geometric subgraph}: Specifies geometric layout constraints (touching, regular grid, separation, etc.) among base blocks.
\par
\textbf{(c) Top geometric subgraph}: Specifies geometric layout constraints among top blocks (when there are multiple).
} \\
\midrule
single-block-stack &
A simple stack: one block sits fully above and is supported by another, forming a minimal stable column or pillar. &
$(\{1\},\{1\})$ &
(a) $\text{supported-by-fully}(\text{top, base})$ \newline
(b) none  \newline
(c) none \\
\midrule
cantilever-with-counterbalance &
Two blocks stacked where the top block overhangs the base, but its center of mass remains safely above the base. &
$(\{1\},\{1\})$ &
(a) $\text{supported-by-partially}(\text{top, base})$ (COM of top within base) \newline
(b) none \newline
(c) none \\
\midrule
two-pillar-single-top-bridge &
Two upright, separated base blocks (pillars) jointly support a single horizontal top block (lintel), which bridges across them. &
$(\{2\},\{1\})$ &
(a) $\text{supported-by-fully}(\text{top, base}_1)$ and $\text{supported-by-fully}(\text{top, base}_2)$ \newline
(b) bases must not touch: $\text{not-touching}(\text{base}_1, \text{base}_2)$. They should be either horizontal-aligned or depth-aligned. \newline
(c) none \\
\midrule
n-pillar-single-top-bridge &
A single bridge block fully supported by $n$ separated pillar blocks below, forming a “wide” bridge. &
$(\{n\},\{1\})$ &
(a) $\text{supported-by-fully}(\text{top, base}_i)$ for all $i=1..n$ \newline
(b) bases must not touch: all pairs $\text{not-touching}(\text{base}_i, \text{base}_j)$, $i\ne j$. The pillars should be arranged in regular-grid-sparse. \newline
(c) none \\
\midrule
single-base-n-pillar-bridge &
A single large base block with $n$ separated vertical pillar blocks supported on top, each independent. &
$(\{1\},\{n\})$ &
(a) $\text{supported-by-fully}(\text{top}_i, \text{base})$ for all $i=1..n$ \newline
(b) none \newline
(c) tops must not touch: all pairs $\text{not-touching}(\text{top}_i, \text{top}_j)$, $i\ne j$. The pillars should be arranged in regular-grid-sparse. \\
\midrule
two-base-single-overhead-pyramid &
Two base blocks, placed touching each other, jointly support a single overhead block centered above. &
$(\{2\},\{1\})$ &
(a) $\text{supported-by-fully}(\text{top, base}_1)$ and $\text{supported-by-fully}(\text{top, base}_2)$ \newline
(b) bases must touch along $x$ or $y$: $\text{touching-along-x}$ or $\text{touching-along-y}$ \newline
(c) none \\
\midrule
n-base-single-overhead-pyramid &
$n$ base blocks form a touching or tightly packed group (typically a regular grid), all together supporting a single overhead block. &
$(\{n\},\{1\})$ &
(a) $\text{supported-by-fully}(\text{top, base}_i)$ for all $i=1..n$ \newline
(b) bases form a regular grid or are all pairwise touching: $\text{regular-grid-compact}$ or all pairwise touching \newline
(c) none \\
\midrule
single-base-n-overhead-pyramid &
A single base supports $n$ top blocks closely packed together (usually in a regular row or grid), like a multi-top step of a pyramid. &
$(\{1\},\{n\})$ &
(a) $\text{supported-by-fully}(\text{top}_i, \text{base})$ for all $i$ \newline
(b) none \newline
(c) tops form a regular grid or are all pairwise touching: $\text{regular-grid-compact}(top_1, ..., top_n)$ or all pairwise touching \\
\midrule
n-base-m-overhead-pyramid &
$n$ base blocks in a compact/touching pattern collectively support $m$ overhead blocks in a similarly compact arrangement; a multi-unit platform or tier. &
$(\{n\},\{m\})$ &
(a) $\text{supported-by-fully/partially}(\text{top}_j, \text{base}_i)$ for each $(i, j)$ where direct support exists \newline
(b) bases: regular grid or all touching, $\text{regular-grid-compact}(base_1,...,base_n)$ \newline
(c) tops: regular grid or all touching, $\text{regular-grid-compact}(top_1,...,top_m)$ \\
\midrule
basic-arc &
Two separated base blocks act as pillars to partially support a top “keystone” block, often at an angle, forming the core of an arch. &
$(\{2\},\{1\})$ &
(a) $\text{supported-by-partially}(\text{top, base}_1)$ and $\text{supported-by-partially}(\text{top, base}_2)$ \newline
(b) bases must not touch: $\text{not-touching}(\text{base}_1,\text{base}_2)$ \newline
(c) none \\
\bottomrule
\caption{Dictionary of local stability patterns. \textbf{Block Counts}: $(\{n\}, \{m\})$ means $n$ base blocks stabilize $m$ top blocks. \textbf{Subgraph Pattern Constraints}: (a) Required supported-by relations between base/top, (b) required geometric relations among base blocks, (c) required geometric relations among top blocks if $m > 1$.}
\label{tab:stability_pattern_dictionary}
\end{longtable}
\section{Training of Diffusion-based Pose Generators for Abstract Relations}
\label{appendix:DM_implementation}

For each distinct abstract relation in our library, we train a dedicated diffusion model to generate object poses that fulfill the specified spatial or stability constraint. Each model is tasked with predicting the denoising direction necessary to recover normalized object poses, ensuring the given structural relation holds.

\paragraph{Model Input and Data Preparation.}
To enable consistent learning, object geometries and poses are preprocessed and normalized: each object's 3D bounding box is scaled relative to the reference container region. The resulting input features comprise shape descriptors (width, height, depth, etc.) and normalized 3D pose coordinates. These, along with a diffusion timestep, form the input tuple for the network.

\paragraph{Modular Encoder Design.}
Our architecture integrates three specialized encoders:
\begin{itemize}
\item \textbf{Shape Encoder:} Implements a two-stage neural network, compressing raw geometry into a 256-dimensional latent representation using SiLU activations.
\item \textbf{Pose Encoder:} With a configuration paralleling the shape encoder, this module transforms the normalized bounding box positions and dimensions into hidden feature space.
\item \textbf{Temporal Encoder:} Timestep information is embedded via a sinusoidal encoder followed by linear and Mish-activated layers, mapping to and from a higher intermediate dimension to facilitate time-aware conditioning.
\end{itemize}

\paragraph{Backbone Architectures.}
The core relational reasoning is handled by one of two network backbones, selected based on relation arity:
\begin{itemize}
\item \textbf{MLP Backbone:} Fixed-arity relations employ a multi-layer perceptron that processes concatenated encodings, mapping directly to noise prediction in pose space. The MLP consists of linear and SiLU layers.
\item \textbf{Transformer Backbone:} Variable-arity relations are addressed with a transformer-based network, which ingests sequences of object features (padded and masked as necessary, with positional encodings) for relations spanning multiple  objects. The transformer output is post-processed and projected to the target pose distribution.
\end{itemize}

\paragraph{Pose Decoding and Reconstruction.}
The pose decoder reverses the encoding process, converting the hidden noise representations produced by the backbone into 3D pose refinements or direct pose predictions as appropriate for each object.

\paragraph{Learning Objective and Inference Procedure.}
The model is trained end-to-end, optimizing mean squared error (L2 loss) between the predicted and true denoising directions at each diffusion step. During inference, a cosine noise schedule over 1500 diffusion steps is applied. For scenarios involving multiple overlapping relations, noise estimates from individual relations are combined—either averaged or weighted—prior to decoding, allowing for joint enforcement of multiple spatial or stability constraints during the generation process.

\section{Baseline Implementation}
\label{appendix:baseline_implementation}

\begin{table}[t]
    \centering
    \caption[Comparison of methods and their user input mode, use of intermediate graphs, and how to search for hidden objects.]{Comparison of methods and their user input mode, use of intermediate graphs, and how to search for hidden objects.}
    \resizebox{\columnwidth}{!}{
    \begin{tabular}{l c c c}
        \hline
        \textbf{Method} & \textbf{User Input Mode} & \textbf{Intermediate Graph} & \textbf{Search for Hidden Objects} \\
        \hline
        Direct VLM Prediction & Language Description & No & VLM informed search \\
        End-to-end Diffusion Model & 2D hand-drawn Sketch & No & N.A. \\
        \hline
        Our Ablation & 2D hand-drawn Sketch & Yes & N.A. \\
        \textbf{Stack It Up (Ours)} & \textbf{2D hand-drawn Sketch} & \textbf{Yes} & \textbf{Stability pattern guided search} \\
        \hline
    \end{tabular}
    }
    \label{tab:method_comparison}
\end{table}

We implemented two baselines and one ablated variant of \model, as summarized in Table~\ref{tab:method_comparison}. The baselines differ in (i) user input modality, (ii) whether they use an intermediate abstract relation graph, and (iii) their ability to infer hidden supporting blocks. We provide implementation details for each below.

\subsection{Direct VLM Prediction}

The Direct VLM Prediction baseline adapts principles from Blox-Net~\citep{goldberg2024blox}, employing a vision-language model (VLM) to generate 3D block arrangements from natural language descriptions. This setup enables us to compare natural language as a specification modality against 2D sketches.

Given a hand-drawn sketch, the pipeline proceeds as follows:

\textbf{1. Scene Description Generation:}
We prompt GPT-4.1 with the 2D sketch to produce a detailed, structured textual description, specifically requesting explicit statements of spatial relations and overall assembly appearance. The instruction emphasizes capturing all necessary details for faithfully reconstructing the depicted 3D structure from language alone. We use the following prompt:
\begin{lstlisting}[language=,caption={Prompt used for scene description generation (Step 1).}]
You a given a front-view 2D rough hand-drawn sketch that illustrate a desired 3D mutli-level stacking structure (in its x-z plane) that are build from rectangle blocks. Generate a detail text description of this sketch, focusing on the relative spatial relations among the objects and the overall appearance, so that someone can generate the 3D structure by specifying the location of the blocks purely based on this textual description.
\end{lstlisting}

\textbf{2. Block Type Selection:}
Using the generated scene description and a catalog of available block types (each with known dimensions), the VLM is asked to:
\begin{itemize}
\item Select a combination of block types and their quantities required to realize the described scene, including both visible and potentially hidden support blocks for stability.
\item Briefly annotate each type's structural role within the assembly.
\end{itemize}
To facilitate this, block types and their dimensions are supplied in a structured dictionary format:
\begin{lstlisting}[language=,caption={Format of candidate block dimensions into the VLM.}]
{
    "type_1_block": [w_1, l_1, h_1], # width (x), length (y), height (z)
    "type_2_block": [w_2, l_2, h_2],
    ...,
    "type_M_block": [w_M, l_M, h_M],
}. 
\end{lstlisting}

Example answer fragment:
\begin{lstlisting}[language=,caption={Example output of block selection via VLM.}]
- 2 x type_3_block: act as the base platform
- 1 x type_7_block: forms the central vertical column
- ...
\end{lstlisting}

Prompt:
\begin{lstlisting}[language=,caption={Prompt used for block selection (Step 2).}]
You are now the structural planner.

Inputs

Textual scene description (from Step 1) - enclosed in <SCENE> ... </SCENE>.
Catalogue of blocks with their exact dimensions - enclosed in <CATALOGUE> ... </CATALOGUE>. The dictionary format is {"type_1_block": [w_1, l_1, h_1], ... } where w = width (x-axis), l = length (y-axis), h = height (z-axis).

Task
Select a set of block types and quantities that can realise the scene while obeying these rules:

- Use only block types listed in the catalogue; no scaling or custom sizes.
- Infer the smallest sufficient number of blocks; you may add hidden support blocks if the scene requires stability.
- Match the relative proportions described in <SCENE>.
- Ignore absolute coordinates-those will be assigned in Step 3.

Output format  (return nothing else)

BLOCK_SELECTION = [
{"type": "type_k_block", "count": N, "role": "one-sentence purpose" },
...
]

Example
BLOCK_SELECTION = [
{ "type": "type_3_block", "count": 2, "role": "forms the ground-level platform" },
{ "type": "type_7_block", "count": 1, "role": "serves as the central vertical column" }
]

Note that "type_0_block" is the base that is not drawn in the sketch. Always select it.

Begin.
\end{lstlisting}  
    
\textbf{3. Block Placement Prediction:}
Finally, the VLM assigns 3D coordinates to every block instance, respecting the following constraints:
\begin{itemize}
\item The entire assembly must fit within a fixed bounding box.
\item Block instances are axis-aligned and individually identified.
\item No two blocks occupy the same space (collisions are disallowed).
\item Support and stability are enforced by allowing the VLM to introduce hidden blocks, as needed.
\item The resulting output is a mapping between each block (with type) and its 3D centroid.
\end{itemize}
The centroid definition, bounding box, and other spatial conventions follow those of our main model for comparability. \newline

Required output format:
\begin{lstlisting}[language=,caption={Pose prediction output format via VLM.}]
{
0: { "type": "type_3_block", "centroid": [-1.2,  0.8, 0.25] },
1: { "type": "type_3_block", "centroid": [ 1.2,  0.8, 0.25] },
2: { "type": "type_7_block", "centroid": [ 0.0,  0.0, 1.75] },
...
}
\end{lstlisting}

Prompt used:
\begin{lstlisting}[language=,caption={Prompt used for block pose generation (Step 3).}]
You are now the placement engine.
Your job is to assign exact 3D positions to every physical block selected in Step 2.

INPUTS

Scene description <SCENE> ... </SCENE>   (optional - use for spatial cues)
Block selection list <SELECTION> ... </SELECTION>   created in Step 2 Example: [ { "type": "type_3_block", "count": 2, "role": "ground-level platform" }, { "type": "type_7_block", "count": 1, "role": "central column" } ]
Block catalogue with dimensions <CATALOGUE> ... </CATALOGUE> Format: { "type_i_block": [w_i, l_i, h_i], ... } w = width (x-axis), l = length (y-axis), h = height (z-axis)
GLOBAL CONSTRAINTS
- The whole assembly must fit inside the bounding box
x \in [-1.5,  1.5]   (width < 3)
y \in [-1.0,  1.0]   (length < 2)
z \in [ 0.0,  5.0]   (height < 5)
- Blocks are axis-aligned; do not rotate them.
- No two blocks may overlap (touching faces/edges is allowed).
- Every block above ground level must rest on, or be supported by, blocks beneath it; hidden support blocks from <SELECTION> may be used.
- Centroids are expressed in the same linear units as the catalogue.

TASK
For every physical block instance:

Assign a unique integer id starting at 0.
Specify its block type (exact key from the catalogue).
Output the centroid coordinates [x, y, z].
OUTPUT FORMAT  (return nothing else)

PLACEMENT = {
0: { "type": "type_3_block", "centroid": [-1.2,  0.8, 0.25] },
1: { "type": "type_3_block", "centroid": [ 1.2,  0.8, 0.25] },
2: { "type": "type_7_block", "centroid": [ 0.0,  0.0, 1.75] },
...
}

Note that "type\_0\_block" is the base that is not drawn in the sketch. Always select it and assigns its centroid to [0, 0, -0.05].

Begin.
\end{lstlisting}

The second baseline follows the end-to-end paradigm of StackGen~\citep{sun2024stackgen}, training a single transformer-based diffusion model to directly predict the full set of 3D block poses conditioned on sketch input, without any explicit intermediate graph representations.

\begin{figure*}[hp]
    \centering
    \includegraphics[width=\linewidth, page=1]{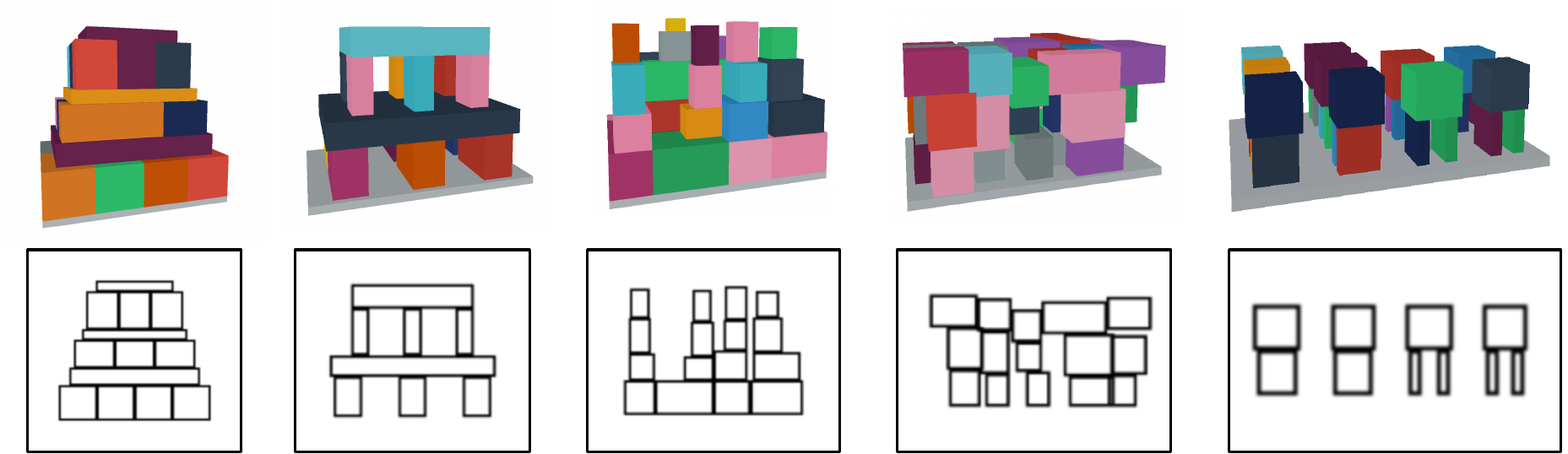}
    \caption[Illustrative examples of sketch conversion.]{\textbf{Illustrative examples of sketch conversion.} For each synthetic 3D block structure (top), we extract the visible blocks from the front-view (x-z plane), and render their outlines as a 2D sketch (bottom). To better resemble human-drawn sketches, we further apply edge dilation and Gaussian blur. These converted sketches serve as input representations for our end-to-end diffusion model baseline.}
    \label{fig:converted_sketch}
\end{figure*}

This baseline involves two steps:
\begin{enumerate}
\item \textbf{Block Type Selection:} A small convolutional neural network (CNN) is trained to predict the set and count of block types required, given the input sketch and candidate block dimensions.
\item \textbf{Block Pose Regression:} A large transformer-based diffusion model generates the 3D poses (centroids) for all selected blocks, conditioned on the sketch and selected types.
\end{enumerate}

The CNN encoder transforms the input sketch into a compact latent embedding by passing it through several convolutional layers, each followed by batch normalization and ReLU activations, and finally by spatial pooling and a linear layer. This embedding is added to each object's feature representation before being processed by the transformer to promote cross-block information sharing and contextualization.

Within the diffusion model, sketch embeddings are broadcast and added to the positional and geometric embeddings of each block, and position encoding is applied. Batch data is padded and appropriately masked to handle varying object counts per sample.

\paragraph{Training and Data:}
Both models are trained on synthetic data generated by composing multiple local stability patterns. However, only the 3D block arrangements and dimensions are available initially. To create paired sketch inputs, we generate a 2D front-view sketch for each 3D structure.

For each sample, we project the 3D arrangement to the x-z (front-view) plane. We retain only the visible blocks, sorting them by their y-coordinates (depth), and filtering out those fully occluded by others. The outlines of visible blocks are rendered onto a blank canvas, and their edges are dilated and blurred to resemble hand-drawn sketches and reduce the domain gap. Padding is added to maintain consistent framing. Figure~\ref{fig:converted_sketch} illustrates representative examples of these generated sketches.

\subsection{Our Ablation: No Hidden Object Prediction}

To assess the importance of hidden support prediction, we ablate the stability-pattern-guided backward graph update in \model. In this variant, the relation graph representing the 3D structure is not expanded with additional hidden support blocks. Instead, when an arrangement generated from the abstract relation graph is found to be unstable in physical simulation, we re-ground the same graph using our compositional diffusion models and attempt pose adjustments using only the initially specified blocks. This allows us to study whether iterative re-sampling alone is sufficient to achieve global stability or if explicit reasoning over hidden supports is necessary.
\section{Hand-drawn Test Cases}\label{appendix:test_cases}

We evaluate all methods on a set of 30 hand-drawn sketches spanning a range of stacking challenges. The complete set of test cases is shown below, across two pages.

\includepdf[pages=-,width=\textwidth]{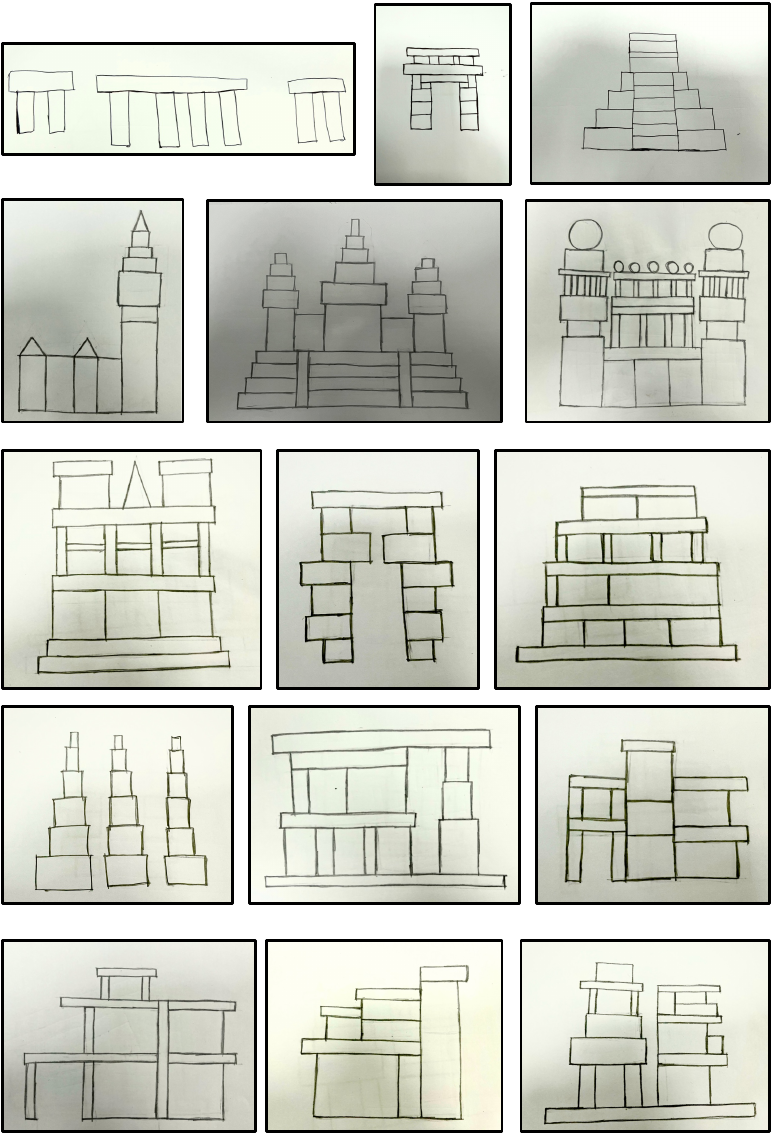}
\includepdf[pages=-,width=\textwidth]{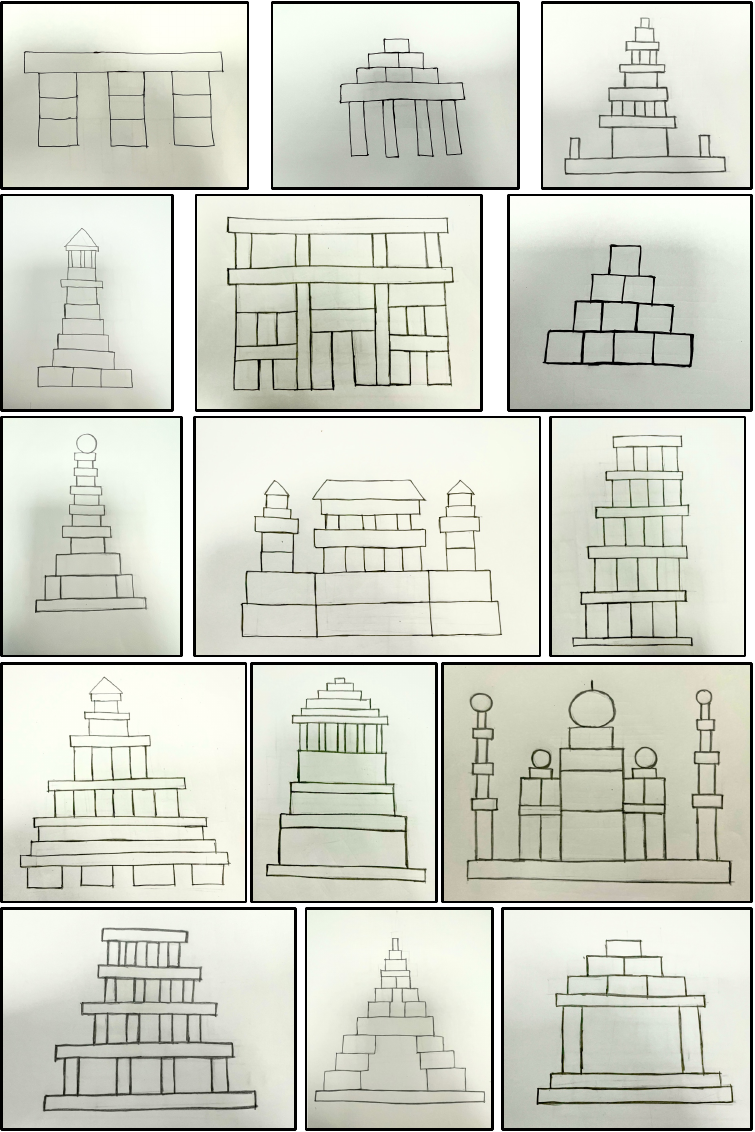}

\captionof{figure}{\textbf{Hand-drawn test cases.} The full set of 30 human-drawn 2D sketches used for evaluation, presented over two pages (15 per page). The test cases cover a variety of multi-level stacking scenarios and structural challenges.}
\label{fig:testcases}
\section{Joint Pose Prediction with Composite Diffusion Scores and ULA}
\label{appendix:joint_ula}

To jointly satisfy multiple abstract relations among objects, we combine the score functions from several trained diffusion models and perform inference using a composite score. This approach enables simultaneous pose generation that respects all specified abstract relations.

\paragraph{Diffusion Reverse Step as Score-Based Sampling.}
A single reverse step of a diffusion model at noise level $t$ updates the input $x_t$ by
\begin{equation*}
x_{t+1} = x_t - \frac{\beta_t}{\sqrt{1 - \bar{\alpha}t}}\epsilon\theta(x_t, t) + \beta_t \xi,\quad \xi \sim \mathcal{N}(0, I),
\end{equation*}
where $\epsilon_\theta$ denotes the neural network’s noise prediction, $\beta_t$ is the step’s noise parameter, and $\xi$ is standard Gaussian noise. Notably, the quantity $\frac{\epsilon_\theta(x_t, t)}{\sqrt{1 - \bar{\alpha}t}}$ is, by denoising score matching theory~\citep{vincent2011connection}, an explicit estimator of the gradient of the log-density for the perturbed data distribution $p_t(x)$ at time $t$:
\begin{equation}
    \nabla_x \log q_t(x) \simeq \frac{\epsilon\theta(x_t, t)}{\sqrt{1 - \bar{\alpha}t}}.
\end{equation}

Thus, the reverse step can be rewritten in Langevin form:
\begin{equation}
x_{t+1} = x_t - \beta_t \nabla_x \log q_t(x) + \beta_t \xi.
\end{equation}

\paragraph{Connection to ULA.}
The unadjusted Langevin algorithm (ULA) samples from $q_t(x)$ according to
\begin{equation}
x_{t+1} = x_t - \eta \nabla_x \log q_t(x) + \sqrt{2 \eta} \xi,\quad \xi \sim \mathcal{N}(0, I),
\end{equation}
for step size $\eta$. If we set $\eta = \beta_t$, this becomes
\begin{equation}
x_{t+1} = x_t - \beta_t \nabla_x \log q_t(x) + \sqrt{2}\beta_t\xi.
\end{equation}
The only difference from the diffusion reverse update is a multiplicative factor of $\sqrt{2}$ in the noise term. Consequently, running the reverse diffusion process is equivalent to a ULA sampler with a reduced noise temperature; one can recover exact ULA sampling by scaling the variance of the added noise by a factor of $2$ at each step, or equivalently scaling the standard deviation by $\sqrt{2}$.

\paragraph{Composing Scores from Multiple Relations.}
When enforcing multiple spatial relations jointly, we aggregate (e.g., sum) the individual score estimates from relevant diffusion models at each step to create a composite score:
\begin{equation}
\nabla_x \log q^t_{\mathrm{prod}}(x) \simeq \underset{r \in \gG}{\sum} w_r \nabla_x \log q^t_{r}(x),
\end{equation}
where $q^t_{r}$ is the noisy distribution associated with relation $r$, and $w_r$ are optional weights. We then perform sampling updates using this composite gradient, applying ULA theory as above.

\paragraph{Implementation in Practice.}
In our experiments, joint ULA sampling is implemented by replacing the noise term in the standard reverse diffusion step with one scaled by $\sqrt{2}$, and substituting the composite score for the individual model score. Alternatively, if using the original (diffusion) noise schedule, the resulting samples correspond to a lower-temperature (less stochastic) variant of the fully tempered ULA trajectory.

\section{Real Robot Execution of Planned Poses}

We conduct our real-robot experiments using a Franka Research 3 (FR3) robotic arm. The robot performs motion planning to reach target gripper poses while avoiding collisions using MoveIt!~\cite{coleman2014reducing}.

We assume that \model's output, $\objectSet = \{o_1, \dots, o_M\}$, is sorted such that lower-level objects have smaller indices. Each object $o_i = (\tau_i, p_i = (x_i, y_i, z_i))$ has a goal pose $\goalPose_i$ with identity orientation:
\begin{equation}
\goalPose_i = \left[
\begin{array}{cccc}
1 & 0 & 0 & x_i \\
0 & 1 & 0 & y_i \\
0 & 0 & 1 & z_i \\
0 & 0 & 0 & 1
\end{array}
\right].
\end{equation}

We assume the initial pose $\initPose_i$ of each object $o_i$ is known. In our setup, we use an L-shaped bracket with a known pose: by aligning the object with the bracket on a flat surface, its pose can be determined. Alternatively, vision-based methods can be used to estimate object poses.

For each object type $\tau'$, we define a relative grasp pose $\graspPose_{\tau'}$ between the gripper frame and the object frame. We also specify a sequence of approach poses $\setTfMatrixPick_{\tau'}$ for picking (\eg, moving the gripper above the object, then descending to align with it). To pick up an object $o_i = (\tau_i, p_i)$, the robot executes the sequence of poses $\initPose_i \graspPose_{\tau_i} \tfMatrix$ for each $\tfMatrix \in \setTfMatrixPick_{\tau_i}$, followed by closing the gripper. For placement, we define pre-placement and post-placement pose sequences, $\setTfMatrixPrePlace_{\tau'}$ and $\setTfMatrixPostPlace$, respectively, which are executed before and after opening the gripper.

To assemble the target configuration $\objectSet$, the robot performs the pick-and-place routine sequentially for each object.

\end{document}